\documentclass{article}

\usepackage[preprint]{neurips_2025}
\usepackage[utf8]{inputenc}
\usepackage[T1]{fontenc}
\usepackage{hyperref}
\usepackage{url}
\usepackage{booktabs}
\usepackage{amsfonts}
\usepackage{amsmath}
\usepackage{amssymb}
\usepackage{nicefrac}
\usepackage{microtype}
\usepackage{xcolor}
\usepackage{graphicx}
\usepackage{algorithm}
\usepackage{algpseudocode}
\usepackage{etoc}
\usepackage{bbm}
\usepackage{titletoc}
\usepackage{circledsteps}
\usepackage[table]{xcolor}
\usepackage{subcaption}
\usepackage{pifont}

\definecolor{deftblue}{RGB}{219,234,254} 

\newcommand{\startappendixtoc}{%
  \startcontents[appendix]
}

\newcommand{\printappendixtoc}{%
  \section*{Appendix Contents}
  \printcontents[appendix]{}{1}{\setcounter{tocdepth}{2}}
}

\algrenewcommand\algorithmicrequire{\textbf{Input:}}
\algrenewcommand\algorithmicensure{\textbf{Output:}}

\title{Expert-Guided Forecast Editing for Time-Series Foundation Models}

\author{%
  Hung Le, Minh Hoang Nguyen, Manh Nguyen, Huu Hiep Nguyen, Dai Do\\
  Deakin Applied Artifical Intelligence Initiative\\
  Deakin University, Australia \\
  \texttt{thai.le@deakin.edu.au}
}

\begin{document}
\maketitle

\begin{abstract}
Time-series foundation models can forecast across heterogeneous domains without task-specific training, but their forecasts are fixed once produced and cannot directly incorporate task-specific expert feedback. We study expert-guided forecast editing: a frozen foundation model generates candidate future trajectories, and an expensive expert evaluator scores them to guide forecast revision. Under a tight query budget, two natural strategies sit at opposite ends: best-of-$N$ purely exploits the foundation model's predictive distribution, while optimization approaches mostly explore the forecast horizon as an unstructured high-dimensional vector. Each extreme is individually sub-optimal. We introduce \textbf{DEFT}, an expert-guided forecast editing framework that balances the two by first exploiting the foundation model's predictive samples in a decomposed trend--seasonal space, then exploring around them via component-wise refinement. DEFT queries the expert only on complete trajectories, then reuses scores for the trend and seasonal components that appeared in the queried recombinations. This lets each expert query provide structured component-level feedback while keeping the foundation model frozen. We compare DEFT against direct search approaches, including best-of-$N$, cross-entropy methods, and Bayesian optimization, under matched expert-query budgets. Across two forecasting benchmarks consisting of 78 datasets, three time-series foundation models, four feedback types, and seven query budgets, DEFT consistently improves the effectiveness of expert guidance. A molecular-dynamics case study further suggests that the same principle extends to more physically grounded feedback, supporting the hypothesis that sparse test-time guidance should be spent balancing prior exploitation with structured exploration.
\end{abstract}

\section{Introduction}

Time-series foundation models (TFMs) have changed the practical form of forecasting. Large forecasting models such as Chronos, TimesFM, Moirai, or Lag-Llama can be trained once and then used on many domains without per-dataset fitting~\citep{rasul2023lag,woo2024unified,das2024decoder, ansari2025chronos}.  Yet, the initial forecast is rarely the final one. Decision makers often bring knowledge that historical data cannot encode, which includes contextual, operational, and domain-specific information that only becomes relevant at prediction time. For example, a retail planner who sees a model predicting a demand surge may know that warehouse capacity is capped at a fixed threshold, making the spike physically impossible to fulfill. In this case, the foundation model produces a reasonable forecast given what it has seen, but the expert holds additional knowledge that demands a targeted revision to the output trajectory.

We study this setting as \emph{expert-guided forecast editing}. A frozen forecasting foundation model first produces an initial forecast for an input series. An editor then proposes candidate future trajectories, and an expert scores a small number of them. The expert may be a human analyst, a simulator, a rule checker, or, in controlled offline experiments, a diagnostic score computed from the held-out future. Because expert calls are costly, the central question is how to spend a small query budget effectively. Two natural strategies sit at opposite ends of an exploit--explore spectrum. \emph{Best-of-$N$}~\citep{snell2024scaling} spends its budget exploiting the foundation model's predictive distribution, asking whether a good trajectory already lies within the model's prior. Optimization-based methods such as the Cross-Entropy Method (CEM,~\citep{rubinstein2004cross}) and Bayesian Optimization (BO,~\citep{shahriari2015taking}) sit at the other end: they use expert feedback to iteratively refine new candidate trajectories, spending the budget exploring regions that the foundation model may not have proposed. Exploitation-heavy methods are ineffective when the prior is biased, while exploration-heavy methods can be wasteful when they ignore a strong prior and search a high-dimensional forecast space with only a small query budget.
We argue that an effective editor should \emph{balance} the two: invest part of the budget in identifying a promising region from the model's predictive distribution, then use the remaining budget to search around it efficiently. Time-series errors typically manifest as level shifts, trend mismatches, or seasonal amplitude changes, so search becomes far more sample-efficient when conducted over these structured components rather than over raw horizon coordinates.

We introduce \textbf{Decomposed Expert-guided ForecasT (DEFT)}, a test-time forecast editor that realizes this exploit-explore balance. DEFT first uses a portion of the query budget to select strong anchor trajectories from the foundation model's predictive distribution, then decomposes candidates into trend and seasonal residual components and searches over their recombinations around the anchor. The expert is queried only on fully reconstructed trajectories, and each score is reused as component-level feedback for the trend and residual proposals that produced it. Hence, a single expert call informs both components while the foundation model remains frozen.
Experimental results across three time-series foundation models, four feedback settings, and 78 time-series datasets show that \textbf{DEFT} improves forecast quality under matched expert-query budgets. A molecular-dynamics case study further suggests that this advantage extends to more physically grounded feedback. The results indicate that sparse test-time guidance is most effective when used to balance TFM-prior exploitation with exploration in component-level space. Our contributions are:

\begin{itemize}
    \item We formulate \emph{expert-guided forecast editing} for frozen TFMs as an exploit–explore budget allocation problem, where, for instance, best-of-N purely exploits the TFM prior and CEM purely explores the horizon space.
    \item We propose \textbf{DEFT}, which balances the two by leveraging strong TFM samples and searching around them over decomposed trend and seasonal components, reusing each expert score as component-level feedback.
   \item We provide a comprehensive evaluation of \textbf{DEFT} against best-of-$N$ variants, random-search, CEM, surrogate-CEM, and BO baselines under the same expert-query budgets, together with ablations and a real-world case study that tests more physically grounded feedback.
\end{itemize}
\section{Related Work}

\textbf{Time-series foundation models.}
Large pretrained forecasting models now provide strong zero-shot forecasts across domains. Chronos tokenizes scaled time-series values and trains language-model architectures for probabilistic forecasting~\citep{ansari2024chronos, ansari2025chronos}. TimesFM pretrains a decoder-only model on large real and synthetic time-series corpora~\citep{das2024decoder}. Lag-Llama and Moirai similarly target general-purpose probabilistic forecasting across datasets, frequencies, and domains~\citep{rasul2023lag,woo2024unified}. Newer models continue this line, e.g., Sundial with flow-matching probabilistic forecasting~\citep{liu2025sundial} and the more compact Moirai 2.0~\citep{liu2025moirai}. These models reduce the need for per-dataset fitting, but their forecasts remain fixed once generated and still show consistent gaps across datasets~\citep{qiao2026s}. Our work is complementary: we keep the foundation model frozen and edit its output forecast using a small number of expert queries to reduce the gaps.

\textbf{Expert feedback and test-time editing for time series.}
Forecasting practice has long studied judgmental adjustment, where human experts override model forecasts using contextual information unavailable in the historical series \citep{lawrence2006judgmental, fildes2009effective}. Our setting differs: the expert does not directly edit forecast values, but provides budgeted scores on complete candidate trajectories, which DEFT uses for structured output-space search. A separate line of time-series test-time adaptation corrects forecasters under temporal drift using delayed observations or online parameter updates~\citep{guo2025online,kim2025battling,im2026cosa}. These methods are not directly applicable to our setting where the future is hidden at decision time, the foundation model remains frozen, and the only feedback is a small number of expert queries before the forecast horizon is observed. To our knowledge, \textit{this is the first work to formulate expert-guided forecast editing for time-series models}, where the forecasting model remains unchanged and scarce expert feedback is used only to revise its output trajectory.

\textbf{Query-limited trajectory search.}
The simplest approach, best-of-$N$, draws $N$ candidates and returns the highest-scoring one; it is widely used for reward-guided selection in language models~\citep{snell2024scaling} but performs no optimization. The Cross-Entropy Method (CEM)~\citep{rubinstein2004cross} iteratively refines a sampling distribution over a forecast horizon, but treats the trajectory as an unstructured vector. Surrogate CEM amortizes feedback using a learned surrogate~\citep{moss2020cross}; Bayesian Optimization~\citep{shahriari2015taking} balances exploration and exploitation via an acquisition function but requires trust regions~\citep{eriksson2019scalable} to scale to high-dimensional horizons. These methods search over the raw, undecomposed horizon. DEFT instead searches over decomposed trend and residual components, querying complete recombined trajectories and reusing each score as component-level feedback.

\textbf{Temporal structure in forecasting.}
Forecasting methods have long used trend, seasonality, and residual structure, from classical decomposition models to neural architectures such as Autoformer, DLinear, and UEC-STD~\citep{wu2021autoformer,zeng2023transformers, nguyen2026reviving}. DEFT uses the same basic temporal bias, but differently: decomposition is not part of a trained forecaster, but rather a test-time editing representation for a frozen foundation model forecast.

\section{Preliminaries}

\subsection{Problem Formulation}
\textbf{Foundation-model forecasting.} Let $x_{1:T}$ denote the observed context and $y_{1:H}$ the unknown future, where each time step may be univariate or multivariate. A time-series foundation model $F_\theta$ produces an initial horizon forecast:
\begin{equation}
          \hat{y}^{0}=F_\theta(x_{1:T}).
\end{equation}
This forecast may be a single trajectory or a predictive distribution represented by samples, quantiles, or other candidate paths. When such distributional outputs are available, we use them as model-derived candidates; otherwise, candidates can be generated by perturbing the point forecast directly in output space. 

\textbf{Expert-guided forecast editing.} An expert feedback scoring function $\operatorname{score}(\cdot)$ provides information about the quality of a complete candidate future trajectory $y$. The feedback may take different forms, including a continuous utility score, a bounded rating, or a preference comparison
between trajectories. The editor $E$ receives the initial forecast $\hat{y}^{0}$, optional model-derived candidates, and a budget $B$ for expert queries. It must return a single edited forecast $\hat{y}$ after querying $\operatorname{score}(\cdot)$ at most $B$ times. Throughout editing, the foundation model remains fixed, i.e., we do not update $\theta$. The foundation model supplies the initial forecast and optionally a prior over plausible futures.

In deployment, the expert may be a human analyst, a downstream simulator, a constraint checker, or a rule-based evaluator. Usually, expert feedback is available only at test time and is expensive. Therefore, the editor cannot ask the expert to evaluate an unrestricted number of candidates; it must use a small number of trajectory-level feedback queries to improve the initial forecast.
\subsection{Query-limited search}
Several search approaches follow naturally from the expert-query setting. We assume throughout that the foundation model $F_\theta$ exposes a predictive distribution over the horizon, either as a set of stochastic samples or as a quantile grid; both forms are supported by all foundation models we consider (Chronos, TimesFM, Moirai). 
\textbf{Best-of-$N$.} The editor constructs a candidate set:
\begin{equation}
\mathcal{Y}=\{y^{(1)},\ldots,y^{(N)}\}
\end{equation}
from the TFM's predictive distribution and returns the highest-scoring trajectory,
\begin{equation}
\hat{y}^{*}=\arg\max_{y\in\mathcal{Y}} \operatorname{score}(y).
\end{equation}
\textbf{Black-box optimization in horizon space.} A second family of baselines actively searches over a posterior proposal distribution refined by expert feedback, rather than relying solely on the TFM prior. For example, \emph{CEM} maintains a diagonal Gaussian proposal,
\begin{equation}
y^{(i)}\sim\mathcal{N}(\mu,\operatorname{diag}(\sigma^2)),
\end{equation}
where $\mu$ can be initialized as $\hat{y}^0$. CEM queries the expert on a budgeted subset sampled from the Gaussian distribution, selects elite set $\mathcal{E}$, and iteratively refits
\begin{equation}
\mu \leftarrow \frac{1}{|\mathcal{E}|}\sum_{y\in\mathcal{E}} y,
\qquad
\sigma^2 \leftarrow \operatorname{Var}_{y\in\mathcal{E}}(y).
\end{equation}
\section{Method}

\begin{figure}[t]
  \centering
  \includegraphics[width=\linewidth]{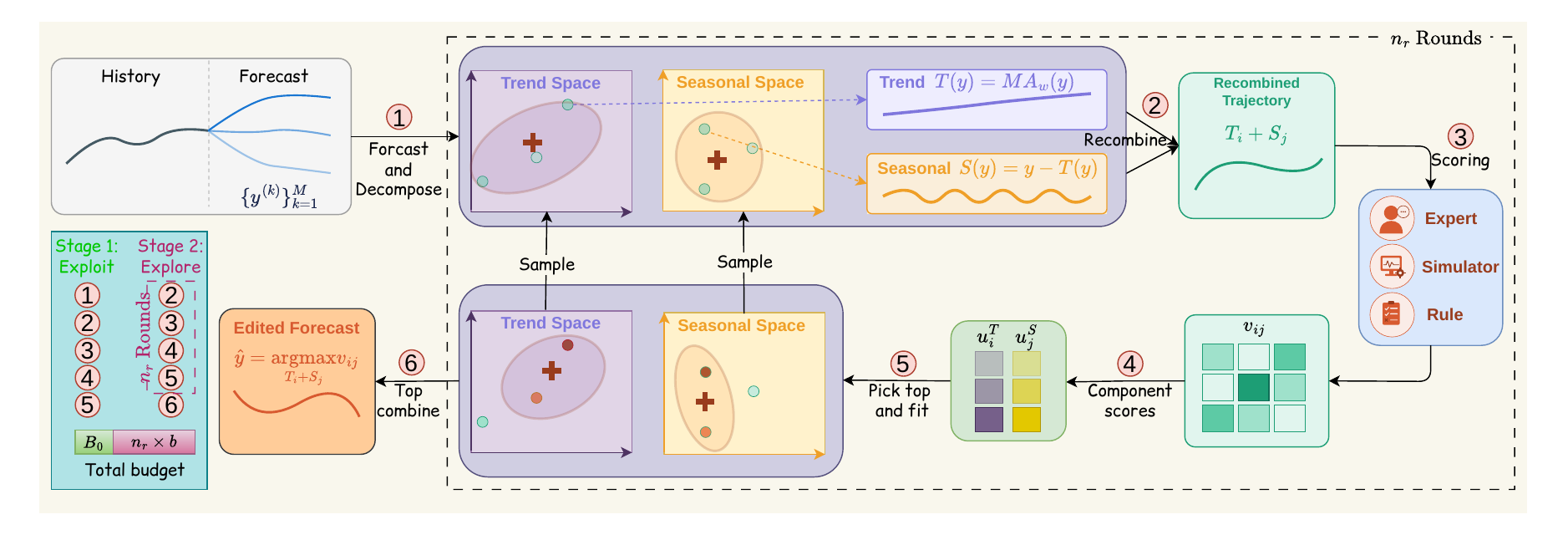}
  \caption{Overview of \textsc{DEFT}. A TFM
  produces initial forecasts, which are decomposed into trend and seasonal components \Circled{1}. \textbf{Stage~1 (Exploit)} scans the model's predictive samples in the decomposed trend--seasonal search space, build
  recombined trajectories $T_i + S_j$ \Circled{2} to query expert for scoring \Circled{3}. It then reuses each score via max-pooling \Circled{4} to
  select elite components and seed component-wise Gaussians \Circled{5}. \textbf{Stage~2
  (Explore)} runs $n_r$ rounds of optimization \Circled{2}--\Circled{5}: it samples new trend and seasonal
  candidates around the seeded means, recombines and queries them, and refits each
  proposal from the elites. The expert may be a human analyst, a downstream
  simulator, or a rule checker. The best trajectory seen across both stages is
  returned as the edited forecast \Circled{6}.}
  \label{fig:deft-overview}
\end{figure}

We propose DEFT (Decomposed Expert-guided ForecasT), a test-time editor that improves a frozen foundation-model forecast using a small number of expert queries. Sec.~\ref{sec:exploit-explore} frames the design choice as a trade-off between exploiting the foundation model's predictive distribution and exploring beyond it. Sec.~\ref{sec:deft} describes DEFT as a principled balance of the two, operating in a decomposed temporal space. Fig. \ref{fig:deft-overview} provides an overview of our approach.

\subsection{An exploit--explore view of the query budget}
\label{sec:exploit-explore}

Given a fixed budget $B$, an editor must decide for each query whether to evaluate a candidate already supported by the TFM prior or to probe a region the prior assigns low mass. Two natural strategies sit at opposite ends of this trade-off. \emph{Best-of-$N$} draws $B$ candidates from the TFM's predictive distribution and returns the highest-scoring one, spending the entire budget exploiting the prior; it cannot recover when no good trajectory lies in the model's support. Search methods such as \emph{CEM} ~\citep{rubinstein2004cross} iteratively refine a Gaussian proposal over the full horizon $\mathbb{R}^H$, spending the budget exploring; with $H$ large and $B$ small, most perturbations move the forecast in directions that do not correspond to real forecast errors. 

We argue an effective editor should \emph{(i)} spend part of the budget exploiting the TFM to identify a strong starting region, \emph{(ii)} explore through a trend--seasonal decomposition that exposes structured error modes such as level, trend, and seasonal-amplitude shifts, and \emph{(iii)} reuse each expert score across the decomposed components it influences.

\subsection{DEFT}
\label{sec:deft}

DEFT realizes the three principles of Sec.~\ref{sec:exploit-explore} by editing trajectories in their trend--seasonal decomposition rather than in raw horizon coordinates. For any candidate $y\in\mathbb{R}^H$ we write
\begin{equation}
y = T(y) + S(y),\qquad T(y) = \operatorname{MA}_w(y),\qquad S(y) = y - T(y),
\label{eq:decomp}
\end{equation}
where $\operatorname{MA}_w$ is a centered moving average of window $w$. $T$ captures the smooth trend and $S$ the higher-frequency seasonal residual. This basis is far lower-dimensional than $\mathbb{R}^H$ for editing purposes and admits independent proposals for the two error modes. DEFT runs in two stages within a total budget $B = B_0 + n_r\times b$, where $B_0$ queries are spent exploiting the TFM prior and the remaining $n_r\times b$ queries refine the search through $n_r$ rounds of $b$ queries each.

 \textbf{Stage 1: exploit the TFM prior in decomposed space.}
From the TFM's prior pool of $B_0$ trajectories $\mathcal{P}=\{y^{(k)}\}_{k=1}^{B_0}$, we form paired component pools
\begin{equation}
\mathcal{P}_T = \{T(y^{(k)})\}_{k=1}^{B_0}, \qquad
\mathcal{P}_S = \{S(y^{(k)})\}_{k=1}^{B_0},
\end{equation}
holding their trend and seasonal parts. We select $K=\lceil\sqrt{B_0}\,\rceil$ representative trend components and $K$ seasonal components (evenly spaced across the pool). The editor forms $B_0$ recombinations of trend and seasonal components: first the $K$ diagonal pairings (trend $i$ with seasonal $i$), which cover every component once, then randomly chosen off-diagonal pairings up to the budget, so each component appears in roughly the same number of recombinations. Balance is important because each component's utility is max-pooled over the recombinations it participates in (Eq.~\ref{eq:utility}); unbalanced participation would bias the elite selection toward over-sampled components. The expert is queried on each \emph{complete} recombined trajectory,
\begin{equation}
v_{ij} = \operatorname{score}(T_i + S_j),\qquad (i,j)\in\mathcal{I}_0,
\end{equation}
and each score is reused as feedback on \emph{both} of its components via max-pooling:
\begin{equation}
u_i^T = \max_{j:\,(i,j)\in\mathcal{I}_0} v_{ij}, \qquad u_j^S = \max_{i:\,(i,j)\in\mathcal{I}_0} v_{ij}.
\label{eq:utility}
\end{equation}
Component-wise elite sets are formed by taking the top-$\rho K$ trend and seasonal components,
\begin{equation}
\mathcal{E}_0^T = \operatorname{TopK}_{\rho K}\{T_i : u_i^T\}, \qquad \mathcal{E}_0^S = \operatorname{TopK}_{\rho K}\{S_j : u_j^S\},
\end{equation}
which initialize component-wise Gaussians $\mathcal{N}(\mu_T,\operatorname{diag}(\sigma_T^2))$ and $\mathcal{N}(\mu_S,\operatorname{diag}(\sigma_S^2))$ with
\begin{equation}
\mu_T = \operatorname{Mean}(\mathcal{E}_0^T),\;\; \mu_S = \operatorname{Mean}(\mathcal{E}_0^S),\;\; \sigma_T = \operatorname{Std}(\mathcal{P}_T),\;\; \sigma_S = \operatorname{Std}(\mathcal{P}_S).
\label{eq:init}
\end{equation}
Importantly, $\sigma_T$ and $\sigma_S$ are initialized from the TFM's own component-wise sample dispersion rather than a hand-tuned hyperparameter, so the search begins at a scale calibrated to model uncertainty. The asymmetry is deliberate: $\mu$ is drawn from the elites to steer the proposal toward the best-scoring region, while $\sigma$ is drawn from the full pool so the search radius reflects model-level uncertainty rather than the tight spread of a few elites. DEFT records the best trajectory seen so far, $\hat y \gets \arg\max_{(i,j)\in\mathcal{I}_0} v_{ij}$.

\textbf{Stage 2: explore via decomposed CEM.}
For $r=1,\dots,n_r$, the editor samples $n_T$ trend candidates and $n_S$ seasonal candidates \emph{independently} from the current component-wise Gaussians,
\begin{equation}
T_i \sim \mathcal{N}(\mu_T,\operatorname{diag}(\sigma_T^2)),\quad i=1,\dots,n_T,
\qquad
S_j \sim \mathcal{N}(\mu_S,\operatorname{diag}(\sigma_S^2)),\quad j=1,\dots,n_S,
\end{equation}
then forms $b$ balanced recombinations $\mathcal{Q}_r = \{T_i+S_j : (i,j)\in\mathcal{I}_r\}$, where $\mathcal{I}_r$ is chosen so that every sampled $T_i$ and every sampled $S_j$ appears in at least one recombination similarly as in Stage 1. The expert is queried on each complete recombined trajectory,
\begin{equation}
v_{ij} = \operatorname{score}(T_i+S_j),\qquad (i,j)\in\mathcal{I}_r,
\end{equation}
and scores are max-pooled to obtain component-level utilities $u_i^T$ and $u_j^S$ as in Eq.~\eqref{eq:utility}. The component proposals are then refit from the elites,
\begin{equation}
\mu_T,\sigma_T \gets \operatorname{Mean},\operatorname{Std}(\mathcal{E}_r^T),\qquad \mu_S,\sigma_S \gets \operatorname{Mean},\operatorname{Std}(\mathcal{E}_r^S),
\end{equation}
where $\mathcal{E}_r^T$ and $\mathcal{E}_r^S$ are the top-$\rho n_T$ trend and top-$\rho n_S$ seasonal elites under $u^T,u^S$. The running best $\hat y$ is updated against every recombination in $\mathcal{Q}_r$, and the loop terminates when the cumulative query count $q = B_0 + rb$ reaches $B$. DEFT returns $\hat y$, the highest-scoring trajectory observed across both stages. Full pseudocode is given in Algorithm~\ref{alg:deft}.

\begin{table}[t]
\centering
\caption{Forecast-editing methods as a factorial over two orthogonal axes: the \emph{search strategy} (exploit the TFM prior, explore with CEM, or both) and the \emph{editing representation} (raw horizon space vs.\ a trend--seasonal
decomposition). ${\star}$ are prior work (Best-of-$N$ and Direct CEM); $\dagger$ are  ablated variants of our method. \textsc{DEFT} is our method combining prior exploitation, exploration, and decomposition.}
\label{tab:factorial}
\begin{tabular}{@{}lcc@{}}
\toprule
 & \multicolumn{2}{c}{\textbf{Editing representation}} \\
\cmidrule(l){2-3}
\textbf{Search strategy} & Raw horizon space & Trend--seasonal decomp. \\
\midrule
Exploit prior ($n_r\!=\!0$) & Best-of-$N^{\star}$       & Pool only$^\dagger$             \\
Explore (CEM)             & Direct CEM$^{\star}$        & No seed pool$^\dagger$            \\
Both                      & CEM $+$ prior$^\dagger$     & \textbf{\textsc{DEFT}}  \\
\bottomrule
\end{tabular}
\end{table}

\textbf{Why DEFT is sample-efficient.}
\label{sec:why-deft}

DEFT improves query efficiency by combining exploitation of the TFM prior with structured search in a decomposed forecast space. In Stage~1, expert queries are spent on recombinations of trend and seasonal components extracted from TFM samples, allowing DEFT to identify useful components already supported by the model while also exploring novel combinations that may not appear in any single forecast sample. In Stage~2, search is performed over separate trend and seasonal proposals rather than directly over the full horizon, focusing exploration on separate components. Furthermore, each expert evaluation of a reconstructed trajectory $T_i+S_j$ contributes feedback to both the trend component $T_i$ and the seasonal component $S_j$ through Eq.~\eqref{eq:utility}, allowing a single query to refine two proposal distributions simultaneously. Together, these design choices extract more information from each expert query than search methods operating directly in the original forecast space. Table~\ref{tab:factorial} summarizes this design space. Existing methods occupy different corners of the exploit--explore and representation axes, whereas \textsc{DEFT} uniquely combines prior exploitation, structured CEM search, and trend--seasonal decomposition.
\section{Experimental Results}

\subsection{Experimental setup}
\label{sec:exp-setup}

\textbf{Baselines.}
We compare DEFT against representative expert-guided editing baselines under matched query budgets. \textit{Zero-shot} uses the median (initial) TFM forecast directly without editing. \textit{Random Search} samples Gaussian perturbations around the median forecast and returns the highest-scoring candidate. \textit{Best-of-$N$ (quantile)} selects the best trajectory from a fixed set of TFM quantile forecasts, while \textit{Best-of-$N$ (sample)} selects from stochastic forecast samples. \textit{Direct CEM} performs cross-entropy search by iteratively sampling candidates, querying the expert, and refitting a Gaussian proposal from elite trajectories \citep{rubinstein2004cross}. \textit{Surrogate CEM} augments CEM with a learned reward surrogate to guide candidate selection under limited expert feedback \citep{moss2020cross}. \textit{TuRBO-1} is a local Bayesian optimization method that fits a Gaussian-process surrogate and searches within an adaptive trust region \citep{eriksson2019scalable}. Details of the baselines are provided in Appendix \ref{app:subsec:baselines}. All methods use the same initial forecast, expert scorer, and expert-query budget.

\textbf{Foundation models.}
We use three state-of-the-art time-series foundation models from independent providers: \textbf{TimesFM 2}~\citep{das2024decoder}, \textbf{Chronos-2}~\citep{ansari2025chronos},  and \textbf{Moirai 2}~\citep{liu2025moirai}. All three expose a predictive distribution as both stochastic samples or a quantile grid, so every baseline draws candidate trajectories from the same model output without altering $\theta$.

\textbf{Expert-feedback modes.}
The expert $\operatorname{score}(\cdot)$ is implemented as an oracle on the held-out future, but exposes one of four feedback interfaces that mimic realistic forms of expert supervision. \textit{Rating-3} and \textit{Rating-5} return bounded discrete ratings by quantizing the continuous score (negative MSE between the forecast and the ground truth) into 3 or 5 equally spaced levels. \textit{Pairwise} returns a binary preference ($\pm1$) indicating whether a candidate is preferred to the foundation model's median forecast under a Bradley--Terry model. \textit{Pairwise-best} instead compares each candidate against the best trajectory found so far by the current search method, providing an adaptive comparison baseline. Implementation details can be found in Appendix \ref{app:subsec:emode}.

\textbf{Query budgets.}
 We sweep $B \in \{2, 4, 8, 16, 32, 64, 128\}$ to characterize each method's behavior from the very-low-budget regime---where best-of-$N$ is competitive---up to budgets large enough for iterative search methods to converge. TuRBO is evaluated only at $B \in \{2, 4, 8, 16\}$: its exact GP surrogate requires $O(n^3)$ time per series as the archive grows to $n = B$ points, making full-benchmark runs at $B \geq 32$ impractical.

\textbf{Metrics.} Following~\citet{ansari2024chronos}, our primary metrics are relative
\textit{MASE} (mean absolute scaled error) and \textit{WQL} (weighted quantile
loss over levels $0.1,0.2,\dots,0.9$), with \textit{MAE} and \textit{MSE} as
secondary metrics (all relative to the zero-shot geometric mean); lower is better for all four.  We additionally report two
\textit{win rates}: the fraction of settings in which a method improves over the zero-shot median (W-ZS) and over random search (W-Rand); higher is better. All metrics use the GluonTS evaluation \citep{JMLR:v21:19-820}. Full definitions are given in Appendix~\ref{app:subsec:metric}.

\textbf{Benchmarks.}
We evaluate on two large-scale time-series benchmark suites. The \textit{Chronos benchmark (ChronosBench)}~\citep{ansari2024chronos} comprises 42 datasets (27 zero-shot, 15 in-domain) spanning horizons from 6 to 56 steps across monthly, quarterly, daily, hourly, and sub-hourly frequencies. \textit{GIFT-Eval}~\citep{aksugift} adds 36 further datasets covering electricity, solar, traffic, weather, ETT, and M4 splits. Together, the two benchmarks yield 78 distinct datasets.  Per benchmark, we evaluate a fixed slice of up to $1{,}000$ series, held constant across all methods, budgets, foundation models, and feedback modes so that every comparison is on identical inputs. Datasets contributing fewer series than their allocation are included in full; larger datasets are randomly subsampled to their cap. This stratification preserves the original Chronos and GIFT-Eval benchmark compositions, keeps the experiment tractable (over half a million expert queries in total), and remains large enough to detect sub-$1\%$ MASE differences between methods.

As a complementary case study, we also evaluate on a molecular-dynamics benchmark (\textit{MD}) from~\citet{le2025accelerating}, where the task is to forecast future atom-position trajectories. This setting allows us to test DEFT with more physically grounded feedback, based on Morse potential violations of predicted molecular configurations, rather than relying solely on synthetic scores.

\textbf{Implementation.}
All experiments are run with three seeds; we report the mean across these seeds, using identical evaluation slices for all methods at every budget. DEFT's budget allocation between the seed-pool scan ($B_0$) and the decomposed CEM rounds ($n_r \times b$) is auto-computed from the total budget $B$ following Algorithm~\ref{alg:deft-budget}. No method-specific hyperparameters are tuned per dataset (see details in Appendix \ref{app:subsec:hp}). We run experiments on a single NVIDIA V100 or H100 GPU and report wall-clock time in Appendix Table~\ref{tab:compute-cost}. Overall, DEFT is nearly as cheap as the Best-of-$N$ family and markedly faster than the remaining search baselines.

\subsection{Main results: budget curves}
\label{sec:main-results}

\begin{table*}[t]
\centering
\caption{Final scores on the \textbf{TimesFM} backbone for the ChronosBench and GIFT-Eval
benchmarks, averaged across budgets,
expert-feedback modes and 3 runs with different seeds. Error metrics are relative to the zero-shot geometric mean ($1.0$ ties zero-shot).
W-ZS / W-Rand are the fraction (\%) of settings beating the zero-shot
median/random search. Arrows indicate whether lower ($\downarrow$) or higher
($\uparrow$) is better. Best per column in \textbf{bold}.}
\label{tab:main-timesfm}
\setlength{\tabcolsep}{4pt}
\resizebox{\textwidth}{!}{%
\begin{tabular}{@{}lcccccccccccc@{}}
\toprule
& \multicolumn{6}{c}{\textbf{ChronosBench}} & \multicolumn{6}{c}{\textbf{GIFT-Eval}} \\
\cmidrule(lr){2-7}\cmidrule(lr){8-13}
Method & MASE\,$\downarrow$ & WQL\,$\downarrow$ & MAE\,$\downarrow$ & MSE\,$\downarrow$ & W-ZS\,$\uparrow$ & W-Rand\,$\uparrow$
       & MASE\,$\downarrow$ & WQL\,$\downarrow$ & MAE\,$\downarrow$ & MSE\,$\downarrow$ & W-ZS\,$\uparrow$ & W-Rand\,$\uparrow$ \\
\midrule
Zero-shot              & 1.000 & 1.000 & 1.000 & 1.000 & --    & 1.8  & 1.000 & 1.000 & 1.000 & 1.000 & --    & 2.8  \\
Random search          & 0.987 & 0.985 & 0.985 & 0.973 & 58.5 & --    & 0.987 & 0.987 & 0.987 & 0.979 & 48.0 & --    \\
Best-of-$N$ (quantile) & 0.887 & 0.902 & 0.902 & 0.865 & 63.0 & 61.0 & 0.884 & 0.873 & 0.873 & 0.804 & 64.3 & 63.1 \\
Best-of-$N$ (sample)   & 1.127 & 1.116 & 1.116 & 1.297 & 15.2 & 9.3  & 1.141 & 1.117 & 1.117 & 1.266 & 10.2 & 7.7  \\
Direct CEM             & 0.986 & 0.981 & 0.981 & 0.960 & 56.3 & 25.9 & 0.985 & 0.985 & 0.985 & 0.976 & 49.4 & 20.9 \\
TuRBO-1                & 0.993 & 0.990 & 0.990 & 0.980 & 49.3 & 25.2 & 0.992 & 0.992 & 0.992 & 0.986 & 46.4 & 27.6 \\
Surrogate CEM          & 0.979 & 0.974 & 0.974 & 0.948 & 60.3 & 39.4 & 0.978 & 0.978 & 0.978 & 0.965 & 52.0 & 31.0 \\
\midrule
\rowcolor{deftblue}
\textsc{DEFT} (ours) & \textbf{0.838} & \textbf{0.865} & \textbf{0.865} & \textbf{0.794} & \textbf{96.1} & \textbf{93.3} & \textbf{0.834} & \textbf{0.826} & \textbf{0.826} & \textbf{0.719} & \textbf{94.0} & \textbf{91.1} \\
\bottomrule
\end{tabular}}
\end{table*}

\begin{table*}[t]
\centering
\caption{Final average scores on the \textbf{Chronos} backbone for the ChronosBench and GIFT-Eval benchmarks, averaged across budgets, expert-feedback modes and  3 runs with different seeds. Error metrics are relative to the zero-shot geometric mean ($1.0$ ties zero-shot).
W-ZS / W-Rand are the fraction (\%) of settings beating the zero-shot
median/random search. Arrows indicate whether lower ($\downarrow$) or higher
($\uparrow$) is better. Best per column in \textbf{bold}.}
\label{tab:final-scores-chronos-nomse}
\setlength{\tabcolsep}{4pt}
\resizebox{\textwidth}{!}{%
\begin{tabular}{@{}lcccccccccccc@{}}
\toprule
& \multicolumn{6}{c}{\textbf{ChronosBench}} & \multicolumn{6}{c}{\textbf{GIFT-Eval}} \\
\cmidrule(lr){2-7}\cmidrule(lr){8-13}
Method & MASE\,$\downarrow$ & WQL\,$\downarrow$ & MAE\,$\downarrow$ & MSE\,$\downarrow$ & W-ZS\,$\uparrow$ & W-Rand\,$\uparrow$
       & MASE\,$\downarrow$ & WQL\,$\downarrow$ & MAE\,$\downarrow$ & MSE\,$\downarrow$ & W-ZS\,$\uparrow$ & W-Rand\,$\uparrow$ \\
\midrule
Zero-shot              & 1.000 & 1.000 & 1.000 & 1.000 & --    & 11.2 & 1.000 & 1.000 & 1.000 & 1.000 & --    & 14.7 \\
Random search          & 0.983 & 0.981 & 0.981 & 0.953 & 60.1 & --    & 0.991 & 0.991 & 0.991 & 0.969 & 41.1 & --    \\
Best-of-$N$ (quantile) & \textbf{0.877} & \textbf{0.863} & \textbf{0.863} & \textbf{0.748} & 75.9 & 70.8 & \textbf{0.927} & 0.922 & 0.922 & 0.870 & 70.4 & 64.5 \\
Best-of-$N$ (sample)   & 1.121 & 1.138 & 1.138 & 1.355 & 17.9 & 12.3 & 1.166 & 1.239 & 1.239 & 1.767 & 11.6 & 9.1  \\
Direct CEM             & 0.977 & 0.975 & 0.975 & 0.952 & 57.8 & 31.6 & 0.986 & 0.985 & 0.985 & 0.966 & 42.8 & 29.4 \\
TuRBO-1                & 0.991 & 0.989 & 0.989 & 0.968 & 54.6 & 29.1 & 0.994 & 0.994 & 0.994 & 0.978 & 41.2 & 31.3 \\
Surrogate CEM          & 0.962 & 0.959 & 0.959 & 0.924 & 63.5 & 47.5 & 0.979 & 0.978 & 0.978 & 0.951 & 46.8 & 38.8 \\
\midrule
\rowcolor{deftblue}
\textsc{DEFT} (ours)   & 0.889 & 0.882 & 0.882 & 0.779 & \textbf{85.0} & \textbf{75.3} & 0.929 & \textbf{0.913} & \textbf{0.913} & \textbf{0.848} & \textbf{79.9} & \textbf{71.7} \\
\bottomrule
\end{tabular}\label{tab:main-chronos-nomse}
}
\end{table*}

\begin{table*}[t]
\centering
\caption{Final average scores on the \textbf{Moirai} backbone for the ChronosBench and GIFT-Eval benchmarks, averaged across budgets,
expert-feedback modes and $3$ runs with different seeds. Error metrics are relative to the zero-shot geometric mean ($1.0$ ties zero-shot). W-ZS / W-Rand are the fraction (\%) of settings beating the zero-shot median / random search. Arrows indicate whether lower
($\downarrow$) or higher ($\uparrow$) is better. Best per column in \textbf{bold}.}
\label{tab:final-scores-moirai}
\setlength{\tabcolsep}{4pt}
\resizebox{\textwidth}{!}{%
\begin{tabular}{@{}lcccccccccccc@{}}
\toprule
& \multicolumn{6}{c}{\textbf{ChronosBench}} & \multicolumn{6}{c}{\textbf{GIFT-Eval}} \\
\cmidrule(lr){2-7}\cmidrule(lr){8-13}
Method & MASE\,$\downarrow$ & WQL\,$\downarrow$ & MAE\,$\downarrow$ & MSE\,$\downarrow$ & W-ZS\,$\uparrow$ & W-Rand\,$\uparrow$
       & MASE\,$\downarrow$ & WQL\,$\downarrow$ & MAE\,$\downarrow$ & MSE\,$\downarrow$ & W-ZS\,$\uparrow$ & W-Rand\,$\uparrow$ \\
\midrule
Zero-shot              & 1.000 & 1.000 & 1.000 & 1.000 & --    & 9.1  & 1.000 & 1.000 & 1.000 & 1.000 & --    & 13.4 \\
Random search          & 0.983 & 0.981 & 0.981 & 0.952 & 62.0 & --    & 0.993 & 0.992 & 0.992 & 0.970 & 44.6 & --    \\
Best-of-$N$ (quantile) & 0.886 & 0.885 & 0.885 & 0.806 & 67.4 & 64.2 & 0.935 & 0.930 & 0.930 & 0.860 & 61.8 & 57.8 \\
Best-of-$N$ (sample)   & 1.031 & 1.041 & 1.041 & 1.094 & 32.0 & 22.9 & 1.058 & 1.063 & 1.063 & 1.090 & 18.0 & 13.5 \\
Direct CEM             & 0.976 & 0.975 & 0.975 & 0.949 & 61.2 & 41.8 & 0.987 & 0.986 & 0.986 & 0.965 & 45.6 & 37.9 \\
TuRBO-1                & 0.991 & 0.990 & 0.990 & 0.970 & 57.8 & 42.7 & 0.995 & 0.994 & 0.994 & 0.977 & 44.0 & 39.9 \\
Surrogate CEM          & 0.962 & 0.959 & 0.959 & 0.922 & 65.6 & 52.0 & 0.980 & 0.979 & 0.979 & 0.949 & 49.9 & 44.0 \\
\midrule
\rowcolor{deftblue}
\textsc{DEFT} (ours) & \textbf{0.849} & \textbf{0.845} & \textbf{0.845} & \textbf{0.741} & \textbf{91.0} & \textbf{85.7} & \textbf{0.906} & \textbf{0.901} & \textbf{0.901} & \textbf{0.815} & \textbf{88.9} & \textbf{84.1} \\
\bottomrule
\end{tabular}\label{tab:main-moirai}
}
\end{table*}

Tables~\ref{tab:main-timesfm}, \ref{tab:main-chronos-nomse}, and \ref{tab:main-moirai} report final scores for TimesFM, Chronos, and Moirai backbones on the ChronosBench and GIFT-Eval benchmarks, averaged over 7 budgets, 4 expert-feedback modes, and 3 seeds. We report the details for each budget and expert mode in Appendix \ref{app:subsec:full_exp}.

On \textbf{TimesFM}, \textsc{DEFT} attains the best value on every error metric and both win rates across both benchmark suites. It lowers MASE to $0.838$ (ChronosBench) and $0.834$ (GIFT-Eval), compared to $0.887$ and $0.884$ for the strongest baseline, best-of-$N$ (quantile). It beats the zero-shot median in $94$--$96\%$ of settings and random search in over $91\%$, roughly 30 points above any competitor.

On \textbf{Chronos},  \textsc{DEFT} delivers the best win rates on both suites ($85.0$/$75.3$ on ChronosBench, $79.9$/$71.7$ on GIFT-Eval) and the lowest WQL, MAE, and MSE on GIFT-Eval. However, best-of-$N$ (quantile) edges it on the raw error metrics for ChronosBench (e.g., $0.877$ vs.\ $0.889$ MASE). We attribute this to Chronos's quantile forecasts being well calibrated, so that simply selecting the best-scoring quantile candidate is already a strong strategy on this backbone and leaves less room for editing to improve the average error. Even so, \textsc{DEFT}'s advantage in win rates shows that its benefit here lies chiefly in consistency across settings rather than in the lowest average error.

On \textbf{Moirai}, \textsc{DEFT} again leads on all error metrics and win rates. It reduces MASE to $0.849$ and $0.906$, well below the strongest baseline ($0.886$ and $0.935$), and is the only method to cut MSE substantially, reaching $0.741$ and $0.815$. It surpasses the zero-shot median in $89$--$91\%$ of settings and random search in $84$--$86\%$, whereas every other editing method stays below $68\%$ against zero-shot.

\subsection{Physically grounded feedback in molecular dynamics}
  \label{sec:md-case-study}

The experts in Section~\ref{sec:main-results} are simulated from the held-out target. However, in real-world applications such as molecular dynamics, matching the ground truth is not always the right
objective. The ground-truth trajectory is only one sample from a stochastic process, so copying it is not the goal; producing a physically plausible trajectory is. We test whether DEFT can effectively edit the forecast of a Li--P--S--Ge solid electrolyte to improve physical plausibility \citep{le2025accelerating}.

\textbf{Setup.} A TFM predicts long future trajectories of atomic positions ($150$ coordinates). The forecast is refined against a physics expert returning the fraction of near-neighbour bonds distorted beyond a fixed tolerance (Morse \emph{bond-violation rate}); datasets, preprocessing, and the violation score follow ~\citet{le2025accelerating}. We sample 1000 history windows of size 192, predict 1000 future steps, and calculate the average violation and MSE. We use Chronos-2 TFM to efficiently handle multivariate inputs and compare \textsc{DEFT} against zero-shot and the strongest competitor, best-of-$N$ (quantile).

\begin{figure}[t]
\centering
\includegraphics[width=\linewidth]{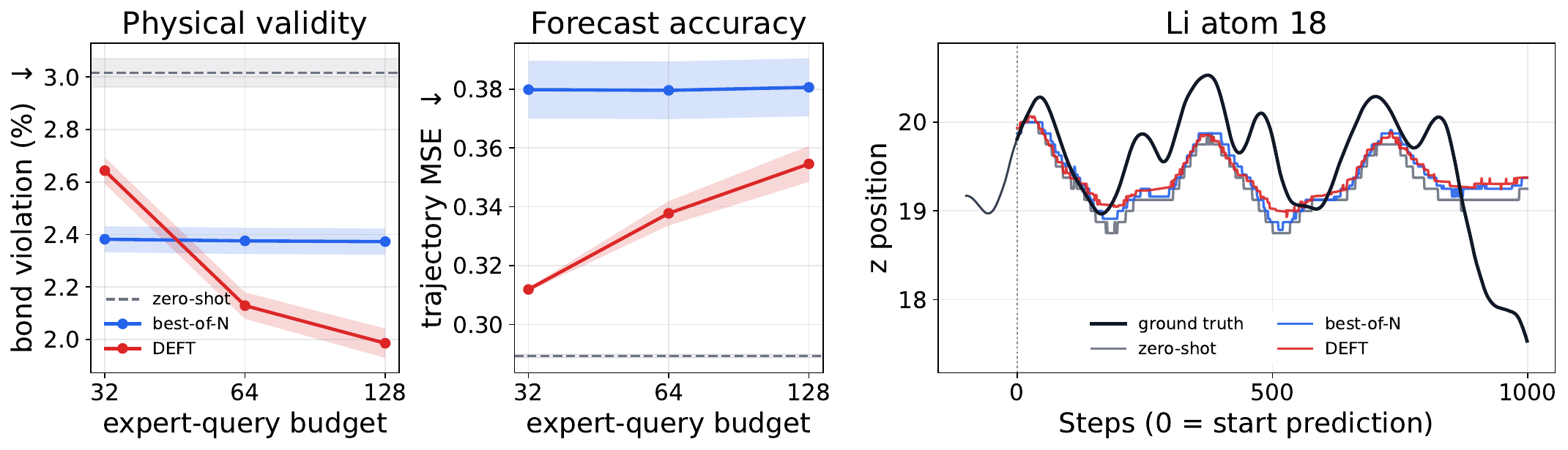}
\caption{MD: Li--P--S--Ge solid electrolyte.
Average bond-violation rate (\textit{left}) and MSE to ground-truth trajectory (\textit{center})  over 3 runs. \emph{Right:} forecast trajectories for a representative
atom (context, ground truth, and edited forecast at $B=128$).}
\label{fig:md}
\end{figure}

\textbf{Results.} Figure~\ref{fig:md} (left) shows that DEFT attains the lowest bond-violation rate and that its advantage \emph{grows with the query budget}: it reduces violation by $13\%$, $30\%$, and $34\%$ relative to the zero-shot forecast at $B=32,64,128$, overtaking best-of-$N$ by $B{=}64$, whereas best-of-$N$ saturates
near $21\%$. Since DEFT and Best-of-$N$ are optimized on bond-violation rate, their MSE to ground truth is increased compared to the zero-shot forecast (centre). Yet, DEFT stays below best-of-$N$ at every budget, making it the Pareto-best refiner. The right panels show the failure mode it corrects, where the zero-shot forecast, despite having the lowest MSE, collapses to a piecewise-constant staircase, which is physically implausible. By contrast, expert-guided search restores plausible atom motion.

\subsection{Ablation study}
\label{sec:ablations}
We isolate the contribution of each DEFT design choice at fixed budgets $B \in \{32,64,128\}$ across three foundation forecasting backbones (Chronos, TimesFM, and Moirai). The definitions of all ablation variants are provided in Appendix~\ref{appendix:ablation-variants}. Figure~\ref{fig:ablation-timesfm-chronos} presents the results for Chronos and TimesFM, while the corresponding Moirai results are deferred to Appendix~\ref{appendix:ablation-results} (Figure~\ref{fig:app:ablation-moirai}). We observe consistent qualitative trends across all three backbones.

\begin{figure}[t]
    \centering
    \includegraphics[width=\linewidth]{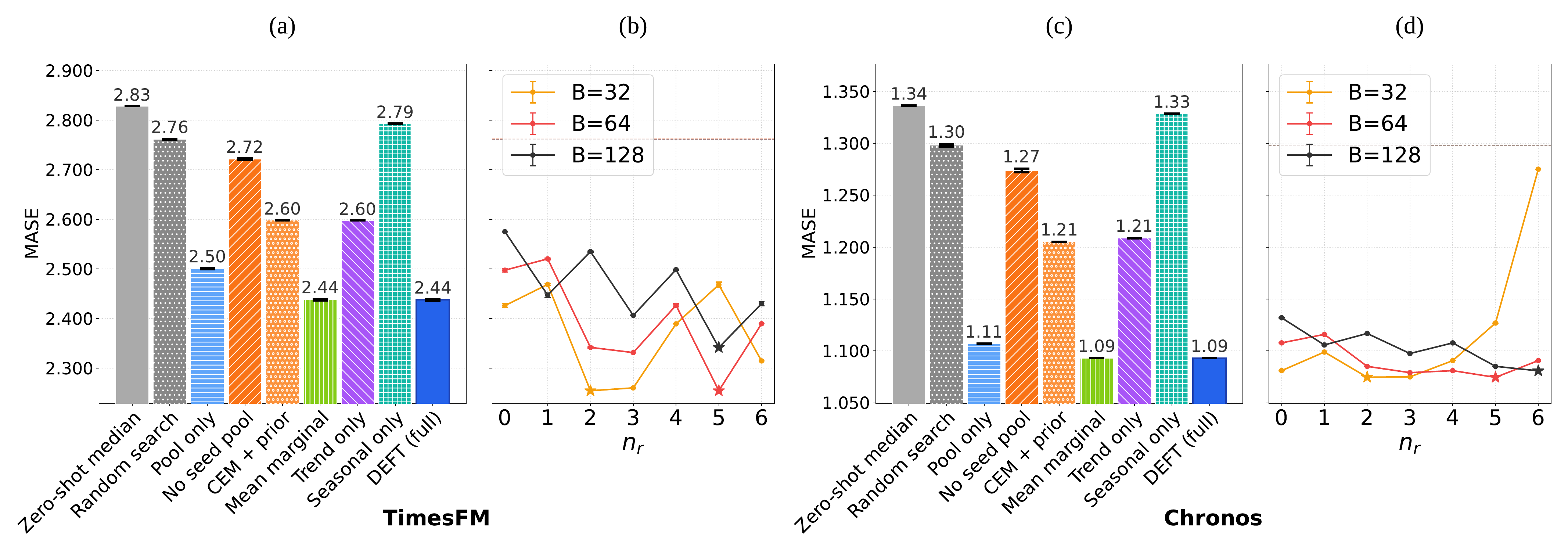}
    \caption{Ablation study of DEFT on the TimesFM and Chronos backbones under inference budgets $B \in \{32,64,128\}$. \textbf{(a)} Mean MASE of each ablation variant on TimesFM. \textbf{(b)} Effect of the number of decomposed CEM refinement rounds ($n_{\text{r}}$) on TimesFM under a fixed evaluation budget, with the prior pool size reduced accordingly; dashed lines denote the random-search baseline and $\bigstar$ indicates the best-performing configuration for each budget. \textbf{(c)--(d)} Corresponding results for Chronos. Panels (a)--(b) share the same TimesFM MASE axis, and panels (c)--(d) share the same Chronos MASE axis. Results for Moirai are provided in Appendix~\ref{appendix:ablation-results}.}
    \label{fig:ablation-timesfm-chronos}
\end{figure}

\textbf{Decomposition vs.\ prior pool.}
Figure~\ref{fig:ablation-timesfm-chronos}(a,c) shows that the quantile-prior seed pool alone explains only part of DEFT's improvement, while trend--residual decomposition provides the majority of the gains. Averaged across budgets, \textit{CEM + prior} reduces MASE from $1.274$ (\textit{No seed pool}) to $1.205$ on Chronos, recovering approximately $38\%$ of the improvement achieved by full DEFT ($1.093$). On TimesFM, it reduces MASE from $2.721$ to $2.598$, accounting for about $43\%$ of the improvement to full DEFT ($2.438$). Likewise, \textit{Pool only} consistently underperforms full DEFT ($1.107$ vs.\ $1.093$ on Chronos; $2.501$ vs.\ $2.438$ on TimesFM), demonstrating that iterative refinement contributes substantially beyond selecting from a high-quality seed pool.

\textbf{Trend vs.\ residual editing.}
Optimizing both decomposed components consistently outperforms optimizing either component alone (Figure~\ref{fig:ablation-timesfm-chronos}(a,c)). Across budgets, \textit{Trend only} remains substantially closer to full DEFT than \textit{Seasonal only} (Chronos: $1.209$ vs.\ $1.329$; TimesFM: $2.598$ vs.\ $2.793$), indicating that trend optimization provides the larger contribution, while jointly optimizing trend and residual components yields the best overall performance.

\textbf{Pool size vs.\ refinement rounds.}
Figure~\ref{fig:ablation-timesfm-chronos}(b,d) examines the trade-off between evaluating the initial seed pool (Stage 1) and iterative CEM refinement (Stage 2). Both TFMs favor additional refinement as the budget increases, with the optimum shifting from $n_r=2$ at $B=32$ to $n_r=5$ at $B=64$. At $B=128$, Chronos continues to benefit from additional refinement, whereas TimesFM peaks at $n_r=5$ and degrades slightly at $n_r=6$, suggesting that TimesFM retains greater value from a larger initial seed pool under large budgets. Nevertheless, performance remains relatively stable around the optimum, indicating that DEFT is not overly sensitive to the exact allocation.

\textbf{Score-reuse via max-pooling.}
Replacing the max-pooled component utility (Eq.~\ref{eq:utility}) with mean pooling yields virtually identical performance on both backbones ($<0.001$ MASE difference: $1.093$ vs.\ $1.093$ on Chronos and $2.438$ vs.\ $2.438$ on TimesFM). This suggests that DEFT's gains stem primarily from reusing candidate evaluations across decomposed components rather than from the specific pooling operator, which motivates our use of max pooling as the default for simplicity.

\begin{figure}[t]
    \centering
    \begin{minipage}[b]{0.32\linewidth}
        \centering
        \includegraphics[width=\linewidth]{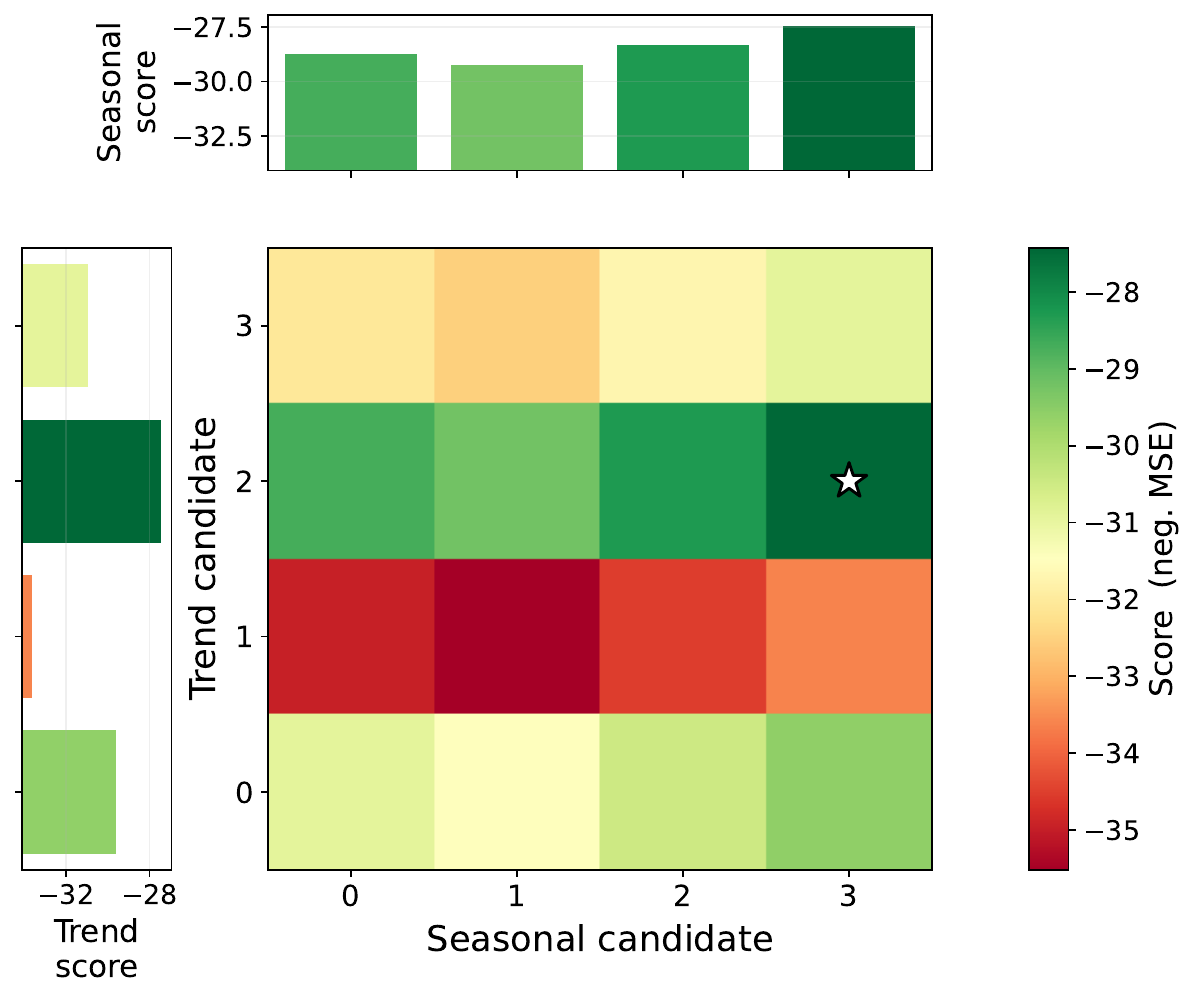}\\
        {\small (a)}
    \end{minipage}
    \hfill
    \begin{minipage}[b]{0.32\linewidth}
        \centering
        \includegraphics[width=\linewidth]{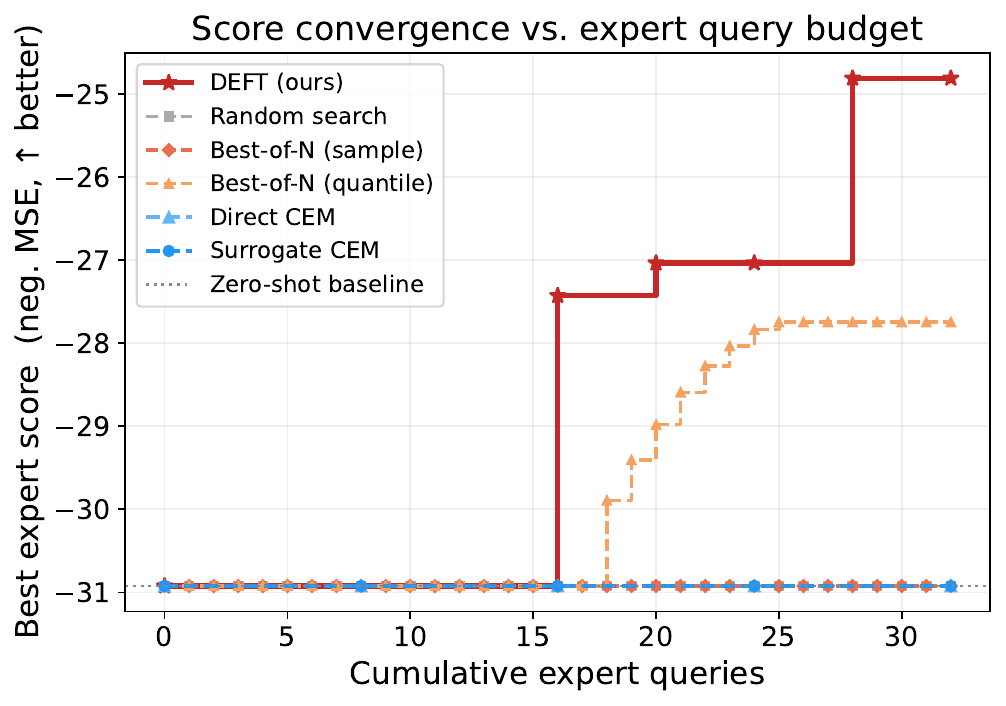}\\
        {\small (b)}
    \end{minipage}
    \hfill
    \begin{minipage}[b]{0.32\linewidth}
        \centering
        \includegraphics[width=\linewidth]{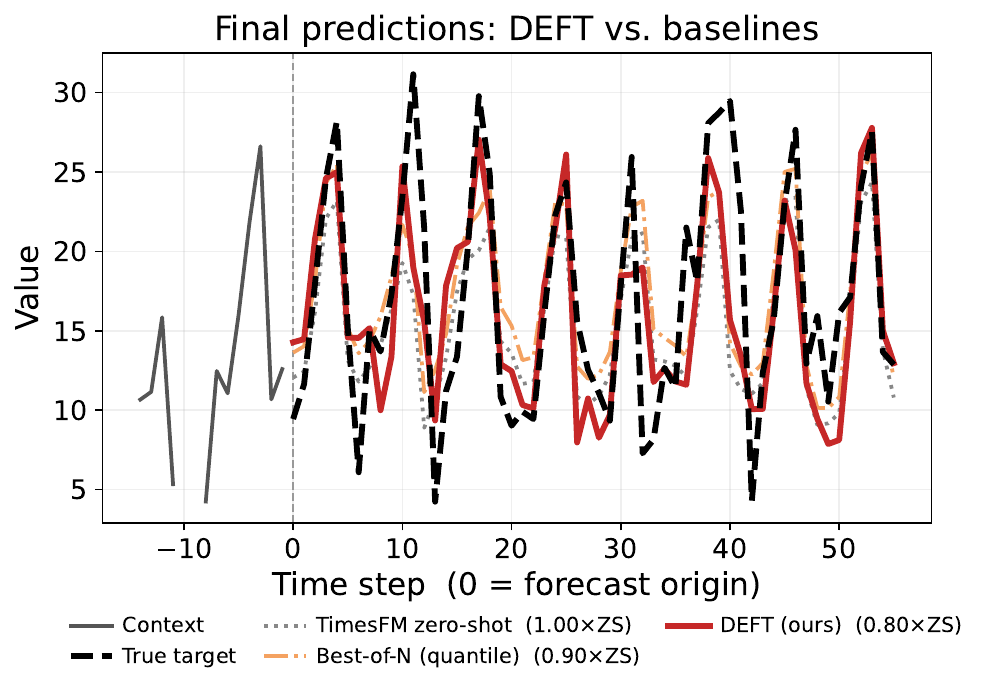}\\
        {\small (c)}
    \end{minipage}
    \caption{DEFT edit on one NN5 series (ChronosBench) with TimesFM, $B=32$. \textbf{(a)} Stage 1 seed-pool scan: cells are measured scores $v_{ij}=\mathrm{score}(T_i+S_j)$ (negative MSE, greener is better); side bars are the max-pooled $u^T_i=\max_j v_{ij}$ and $u^S_j=\max_i v_{ij}$ (Eq.~\ref{eq:utility}); star marks the best pair. \textbf{(b)} Best score vs.\ cumulative queries. \textbf{(c)} Edited forecast; legend gives MSE relative to zero-shot.}
    \label{fig:analysis}
\end{figure}

\subsection{Model Analysis}
In this section, we further examine \emph{how} DEFT edits a single forecast. We analyze one representative edit end-to-end: how the query budget is spent and what the returned forecast looks like.

\textbf{The trend component drives forecast quality.} In panel~(a), each cell is a measured score of a pair $T_i+S_j$, and the side bars are the max-pooled component scores of Equation \ref{eq:utility}. Since these are maxima over measured scores, each bar is simply the best cell in its row or column. The trend scores span $6.2$ against only $1.8$ for the seasonal ones, so the trend drives quality, matching the Trend only versus Seasonal only ablation in Section \ref{sec:ablations}. 

\textbf{Query budget.} Panel~(b) shows DEFT uses the budget better than baselines. Its Stage 1 scan alone, at 16 queries, already beats what Best-of-$N$ (quantile) reaches with all 32 ($-27.43$ against $-27.75$). After 16 queries, DEFT's Stage 2 rounds push it further to $-24.81$, achieving the best results. 

\textbf{Comparison over the whole prediction window.} DEFT reaches the lowest error in panel~(c): $0.80\times$ the zero-shot MSE against $0.90\times$ for Best-of-$N$ (quantile), so $20\%$ below zero-shot and $11\%$ below the best baseline. The gain is not spread evenly, by design: the expert scores a whole trajectory, so DEFT is rewarded for fixing a few large errors, not for staying close everywhere. The zero-shot forecast is flat  (standard deviation $4.51$ against $7.17$, for the target) and DEFT restores much of the swing ($5.71$). That helps most at the peaks zero-shot misses, at the cost of small overshoots elsewhere, and DEFT beats zero-shot at $62\%$ of time steps.

\section{Conclusion}
We introduced \textsc{DEFT}, a test-time editor that revises the forecast of a
frozen time-series foundation model using a small budget of expert queries. We
framed expert-guided forecast editing as an exploit--explore budget-allocation
problem, and \textsc{DEFT} balances the two ends by leveraging strong samples
from the model's predictive distribution and searching around them in a decomposed
trend--seasonal space. Across three foundation-model backbones, four expert-feedback modes, and 78 datasets under
matched query budgets, \textsc{DEFT} consistently improves forecast quality over
best-of-$N$, random search, CEM, surrogate CEM, and Bayesian optimization. These
results support our central hypothesis: under a tight query budget, sparse
test-time guidance is better spent balancing prior exploitation with structured
exploration than on arbitrary horizon-space perturbations.

\section*{Limitations}

Our main experiments use an expert derived from the held-out future. This gives a
controlled and reproducible signal, but does not capture the noise,
miscalibration, or contextual knowledge of a real human analyst; evaluating
\textsc{DEFT} with human experts and hard-constraint checkers remains future work.
The trend--seasonal decomposition also assumes forecast errors are well described
by level, trend, and seasonal-amplitude modes, which may not hold for series
dominated by regime shifts or irregular structure. Finally, we keep the foundation
model frozen throughout; combining test-time editing with lightweight adaptation of the model itself may yield further gains.


\bibliographystyle{plainnat}
\bibliography{references}

\clearpage
\appendix

\startappendixtoc
\printappendixtoc

\clearpage
\section{Implementation Details}

\subsection{Baselines}\label{app:subsec:baselines}

We provide details for all the methods studied in our paper as follows: 

\begin{itemize}
    \item \textbf{Random Search} perturbs the foundation-model median with i.i.d. Gaussian noise to draw $B$ candidate trajectories and returns the one the expert scores best. It has no iterative refinement and serves as the simplest search baseline (details in Algorithm~\ref{alg:random-search}).
    \item \textbf{Best-of-$N$ Quantile}  selects the highest-scoring trajectory from a fixed set of $N$ quantile forecasts produced by the foundation model. Specifically, we query the model's quantile head at a symmetric grid of $N$ quantile levels centered at the median (always including $q=0.5$), producing $N$ deterministic full-horizon trajectories that are scored by the expert (details in Algorithm~\ref{alg:bon-quantile}). 
    \item \textbf{Best-of-$N$ Sample} selects the highest-scoring trajectory from $N$ stochastic forecast samples. For models with native sampling, trajectories are drawn directly; for quantile-only models (e.g., Chronos-2), samples are approximated using per-timestep inverse-CDF sampling over a dense quantile grid, while always including the median forecast (details in Algorithm~\ref{alg:bon-sample}).
    \item \textbf{TuRBO-1} runs local Bayesian optimisation in horizon space: it fits a Gaussian-process surrogate to all past expert queries and each round samples a batch of candidates inside a trust region around the current best, enlarging the region after improving rounds and shrinking it after stalled ones. Because the Gaussian process scales cubically with the number of queries, it is only run at small budgets ($B<32$) (details in Algorithm~\ref{alg:turbo}).
    \item \textbf{Direct CEM} runs the cross-entropy method directly in horizon space: at each round it samples candidates from a Gaussian proposal, queries the expert on a budgeted subset, and re-fits the proposal to the queried elites. Every proposal update is driven purely by expert scores (details in Algorithm~\ref{alg:direct-cem}).
    \item \textbf{Surrogate CEM} augments the same CEM loop with a ridge-regression surrogate fit on all past expert queries. The surrogate ranks unqueried candidates so each round spends its budget on a mix of high-predicted and archive-far candidates, amortising expert calls (details in Algorithm~\ref{alg:surrogate-cem}).
    \item \textbf{Decomposed Expert-Guided Forecast Editing (DEFT - Ours).} DEFT decomposes the median into a trend and a seasonal component and searches them separately. A free pool of foundation-model quantile trajectories seeds per-component proposals, and each round scores full recombinations $T_i + S_j$ while attributing quality back to individual components, so a single expert query informs both proposals (details in Algorithm~\ref{alg:deft}).
\end{itemize}

\subsection{Hyperparameters}\label{app:subsec:hp}

All methods share a matched-budget protocol: for budget $B$ each editor issues at most $B$ expert queries per series, starts from the same initial forecast $\hat{y}^0$, and is scored by the same expert. We use one global configuration for all $78$ datasets, three backbones, and four feedback modes.

\paragraph{Foundation Models.} The exact checkpoints, loader packages, and inference settings are listed in Table~\ref{tab:tfm-versions}. Chronos-2 is run in its native multivariate mode and exposes a full quantile output, whereas TimesFM and Moirai are applied univariately to each coordinate and expose only the $[0.1,0.9]$ quantile grid. In all cases, the model weights are held constant throughout.

\paragraph{Baselines.} The best-of-$N$ methods use $N=B$ candidates scored in a single pass, and \textsc{Random Search} draws $B$ Gaussian perturbations of $\hat{y}^0$. The key knobs for the CEM-family baselines (\textsc{Direct}/\textsc{Surrogate CEM}) are the queries per iteration $b=\max(\lfloor B/4\rfloor,2)$ and the elite fraction $\rho=0.2$ used to refit a diagonal Gaussian each iteration. \textsc{TuRBO-1}, being $\mathcal{O}(n^3)$ in the query count, is run only at $B<32$.

\paragraph{DEFT.} The two key design choices are the trend--seasonal decomposition and the exploit/explore budget split. DEFT edits with $T(y)=\mathrm{MA}_w(y)$, where the window $w$ is inferred per series from the dominant autocorrelation period (fallback $\max(3,\lfloor H/4\rfloor)$, clamped to $[2,H]$) and is the only per-series quantity. The budget splits as $B=B_0+n_r\,b$: $B_0$ seed-pool queries and $n_r$ refinement rounds of $b$, computed based on Algo. \ref{alg:deft-budget} given query budget $B$. Intuitively, DEFT aims to split its budget evenly—half to a one-shot scan of a diverse seed pool (breadth), half to iterative CEM refinement (depth). At small budgets, though, a single round cannot reliably fit its additive trend+seasonal model, so the allocation defers refinement and spends the budget almost entirely on the seed-pool scan.

We report the resulting values in Table~\ref{tab:deft-alloc}. DEFT reuses $\rho=0.2$ for the component elites but, crucially, initialises the proposal scales from the model's own sample spread ($\sigma_T=\mathrm{Std}(\mathcal{P}_T)$, $\sigma_S=\mathrm{Std}(\mathcal{P}_S)$) rather than a hand-set value, and splits each round evenly between trend and seasonal candidates ($n_T=n_S=b/2$).

\begin{table}[t]
\centering
\caption{Time-series foundation models used as frozen forecasters. ``Multiv.'' indicates whether the checkpoint is run in native multivariate mode.}
\label{tab:tfm-versions}
\setlength{\tabcolsep}{5pt}
\renewcommand{\arraystretch}{1.25}
\begin{tabular}{@{}lllcc@{}}
\toprule
\textbf{Foundation model} & \textbf{Checkpoint (HF id)} & \textbf{Multiv.} & \textbf{Quantile grid} \\
\midrule
Chronos-2  & \texttt{amazon/chronos-2}  & \ding{51} & full \\
TimesFM-2.0 (500M) & \texttt{google/timesfm-2.0-500m-pytorch} & \ding{55} & $[0.1,0.9]$ \\
Moirai-2.0 (R, small) & \texttt{Salesforce/moirai-2.0-R-small} & \ding{55} & $[0.1,0.9]$ \\
\bottomrule
\end{tabular}
\end{table}

\begin{table}[t]
\centering
\caption{DEFT budget allocation from total budget $B=B_0+n_r\,b$; $n_T/n_S$ are the
trend/seasonal candidates per round.}
\label{tab:deft-alloc}
\begin{tabular}{@{}ccccccc@{}}
\toprule
$B$ & $B_0$ & $b$ & $n_r$ & $n_T$ & $n_S$ \\
\midrule
$2$   & $2$  & $4$  & $0$ & --  & --  \\
$4$   & $4$  & $4$  & $0$ & --  & --  \\
$8$   & $4$  & $4$  & $1$ & $2$ & $2$ \\
$16$  & $12$ & $4$  & $1$ & $2$ & $2$ \\
$32$  & $16$ & $4$  & $4$ & $2$ & $2$ \\
$64$  & $32$ & $8$  & $4$ & $4$ & $4$ \\
$128$ & $64$ & $16$ & $4$ & $8$ & $8$ \\
\bottomrule
\end{tabular}
\end{table}

\subsection{Expert-feedback modes}\label{app:subsec:emode}

In this section, we present the formulation for each expert-feedback mode.
Let $J(y)=\tfrac{1}{H}\lVert y-y^\star\rVert_2^2$ be the hidden objective (MSE to the unseen future $y^\star$, lower is better) and $J_{\mathrm{ref}}$ a reference scale, by default the foundation-model median MSE $J(\hat{y}^{0})$. The four feedback modes map $J$ to progressively coarser signals, summarised in Table~\ref{tab:expert-modes}: \emph{rating} quantises a bounded quality score to $N$ equally spaced levels; \emph{pairwise} returns a $\pm1$ preference against the fixed median $\hat{y}^{0}$; and \emph{pairwise-best} compares instead against an anchor that advances to the best objective found so far, making later batches harder to satisfy.

\subsection{Metrics}\label{app:subsec:metric}

Following prior work~\citep{ansari2024chronos}, we evaluate forecasting performance using Mean Absolute Scaled Error (MASE), Weighted Quantile Loss (WQL), Mean Absolute Error (MAE), Mean Squared Error (MSE), and pairwise win rate. Let $y_h,\; h=1,\ldots,H$ denote the ground-truth target values over the forecast horizon of length $H$, and $\hat{y}_h,\; h=1,\ldots,H$ the corresponding predicted values. Let $x_t,\; t=1,\ldots,T$ denote the observed context of length $T$, where $m$ is the seasonal period used for scaling MASE. For WQL, $\hat{y}^{(q)}$ denotes the prediction at quantile level $q\in\mathcal{Q}$, where $\mathcal{Q}=\{0.1,0.2,\ldots,0.9\}$. Win rates are computed over the set of evaluated time series $\mathcal{S}$.

\paragraph{Mean Absolute Scaled Error (MASE)}
\begin{equation}
\mathrm{MASE}=
\frac{\frac{1}{H}\sum_{h=1}^{H}|y_h-\hat y_h|}
{\frac{1}{T-m}\sum_{t=m+1}^{T}|x_t-x_{t-m}|}.
\end{equation}

\paragraph{Weighted Quantile Loss (WQL)}
\begin{equation}
\mathrm{WQL}=
\frac{\sum_{q\in\mathcal Q}\sum_{h=1}^{H}
\Lambda_q(y_h,\hat y_h^{(q)})}
{\sum_{h=1}^{H}|y_h|},
\end{equation}
where
\begin{equation}
\Lambda_q(y,\hat y)=
\begin{cases}
q(y-\hat y), & y\ge\hat y,\\
(1-q)(\hat y-y), & y<\hat y.
\end{cases}
\end{equation}

\paragraph{Mean Absolute Error (MAE) and Mean Squared Error (MSE)}
\begin{equation}
\mathrm{MAE}=
\frac{1}{H}\sum_{h=1}^{H}|y_h-\hat y_h|,
\qquad
\mathrm{MSE}=
\frac{1}{H}\sum_{h=1}^{H}(y_h-\hat y_h)^2.
\end{equation}

\paragraph{Win Rate}
\begin{equation}
    \text{W-ZS} = \frac{1}{|\mathcal{S}|}\sum_{i\in S}\mathbbm{1}\!\left[\mathrm{MASE}_i^{\text{method}} < \mathrm{MASE}_i^{\text{zs\_median}}\right],
\end{equation}
\begin{equation}
    \text{W-Rand} = \frac{1}{|\mathcal{S}|}\sum_{i\in S}\mathbbm{1}\!\left[\mathrm{MASE}_i^{\text{method}} < \mathrm{MASE}_i^{\text{random\_search}}\right].
\end{equation}

\begin{table}[t]
\centering
\caption{Expert-feedback modes. Each score already includes the noise level $\eta\ge0$: additive Gaussian for the cardinal modes and a Bradley--Terry temperature for the pairwise modes.}
\label{tab:expert-modes}
\begin{tabular}{@{}llc@{}}
\toprule
Mode & Signal & Score \\
\midrule
Rating ($N$) & $N$ discrete levels & $\operatorname{score}(y) = 1 + \operatorname{round}\!\Big(\dfrac{\tilde{s}(y)-1}{9}(N-1)\Big) + \varepsilon$ \\[10pt]
Pairwise & $\pm1$ vs.\ fixed median &
$\operatorname{score}(y) = \begin{cases} +1 & \text{w.p. } \sigma\!\big(m(y)/\tau\big) \\ -1 & \text{otherwise} \end{cases}$ \\[12pt]
Pairwise-best & $\pm1$ vs.\ advancing anchor &
$\operatorname{score}(y) = \begin{cases} +1 & \text{w.p. } \sigma\!\big((J_{\mathrm{anchor}}-J(y))/\tau\big) \\ -1 & \text{otherwise} \end{cases}$ \\
\midrule
\multicolumn{3}{@{}p{\linewidth}@{}}{%
where $\varepsilon\sim\mathcal{N}(0,\eta^2)$, $\tilde{s}(y)=1+\dfrac{9}{1+J(y)/J_{\mathrm{ref}}}\in(1,10]$, $m(y)=J(\hat{y}^{0})-J(y)$, $\tau=\eta J_{\mathrm{ref}}$, $\sigma(\cdot)$ is the logistic function, and $J_{\mathrm{anchor}}\!\gets\!\min(J_{\mathrm{anchor}},\min_i J(y_i))$ advances after each batch. At $\eta=0$ the pairwise scores reduce to $\operatorname{sign}(\cdot)$.} \\
\bottomrule
\end{tabular}
\end{table}

\pagebreak

\section{Full Experimental Results}\label{app:subsec:full_exp}

In this section, we report detailed budget curves for each metric, tracing how every method scales with budget $B$. Overall, we can see that methods fall into three regimes: selection-based baselines that improve quickly but plateau, optimization-based baselines (the CEM family and TuRBO-1) that need large budgets before making meaningful progress, and \textsc{DEFT}, which is both strong at small budgets and continues to scale.

Among the baselines, Best-of-$N$ (sample) barely moves off the zero-shot line throughout, as sampled trajectories are noisy candidates for selection. The CEM family (Direct CEM, Surrogate CEM) and TuRBO-1 improve only slowly at small budgets: Surrogate CEM, the strongest of these, does not approach competitive error until $B\!\geq\!32$ and still trails at $B\!=\!128$. Best-of-$N$ (quantile), by contrast, is remarkably strong at intermediate budgets. It often drops sharply and by $B\!=\!8$--$16$. This is expected because when the backbone's quantile forecasts are well behaved, a handful of quantile candidates span most of the useful predictive range. Hence, a modest $N$ suffices to pick a near-optimal one. Beyond this point, however, it \emph{saturates} and additional budget yields almost no further gain (its MASE/MSE curves flatten from $B\!=\!16$ onward), because selection can only choose among the fixed pool of quantile candidates and cannot construct better forecasts outside it.

\textsc{DEFT} exhibits the opposite, more desirable scaling. It starts strong even at $B\!=\!2, 4$ and continues to improve steadily as the budget grows, overtaking Best-of-$N$ (quantile) on MSE at large $B$ and reaching the lowest error at $B\!=\!128$, since editing can refine forecasts rather than merely select among them. Crucially, \textsc{DEFT} is far more budget-efficient than the CEM family, matching or beating their best large-budget error with an order of magnitude less budget. This efficiency is clearest in the win-rate panels, where \textsc{DEFT} always exceeds against random search from the smallest budgets and stays near the top throughout, whereas the CEM methods only catch up once $B$ becomes large. Below are the report details. 

\begin{figure*}[h]
    \centering
    \includegraphics[width=\textwidth]{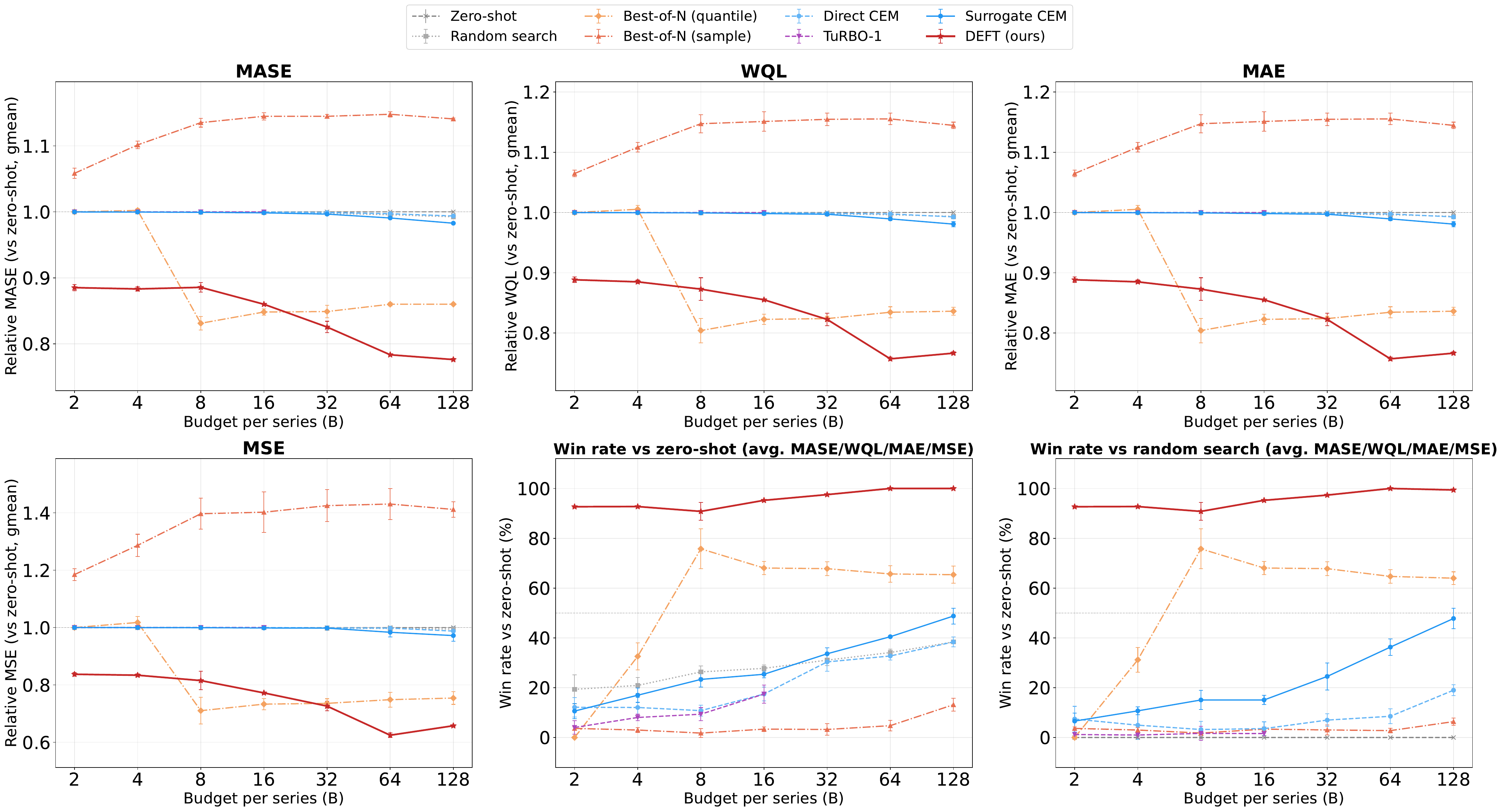}
    \caption{Performance across expert-query budgets on the Chronos Benchmark using the TimesFM foundation model with 3-Level Rating expert feedback.}
    \label{fig:budget-chronosbm-timesfm-rating3-noise0}
\end{figure*}

\begin{figure*}[h]
    \centering
    \includegraphics[width=\textwidth]{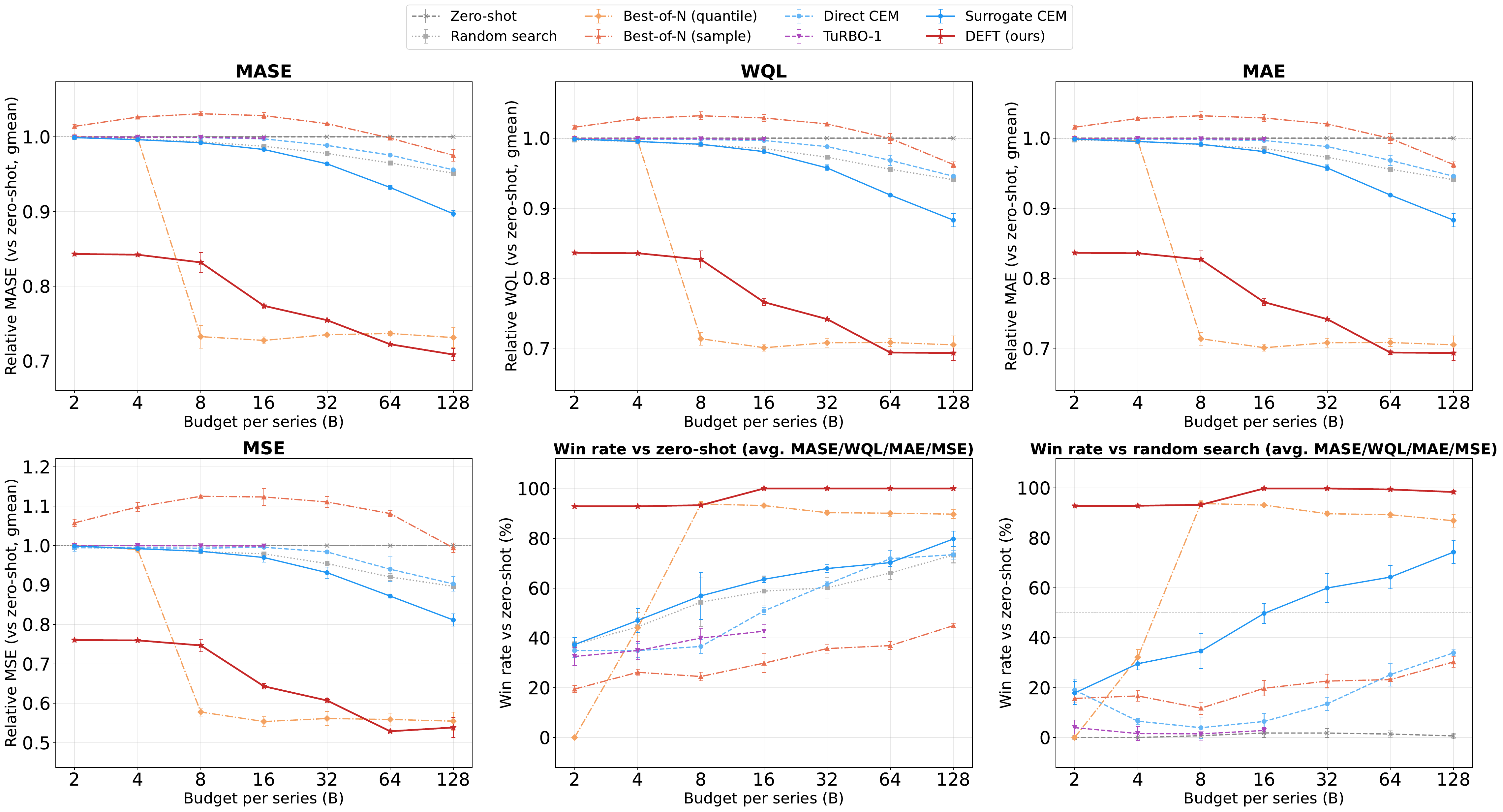}
    \caption{Performance across expert-query budgets on the Chronos Benchmark using the TimesFM foundation model with 5-Level Rating expert feedback.}
    \label{fig:budget-chronosbm-timesfm-rating5-noise0}
\end{figure*}

\begin{figure*}[h]
    \centering
    \includegraphics[width=\textwidth]{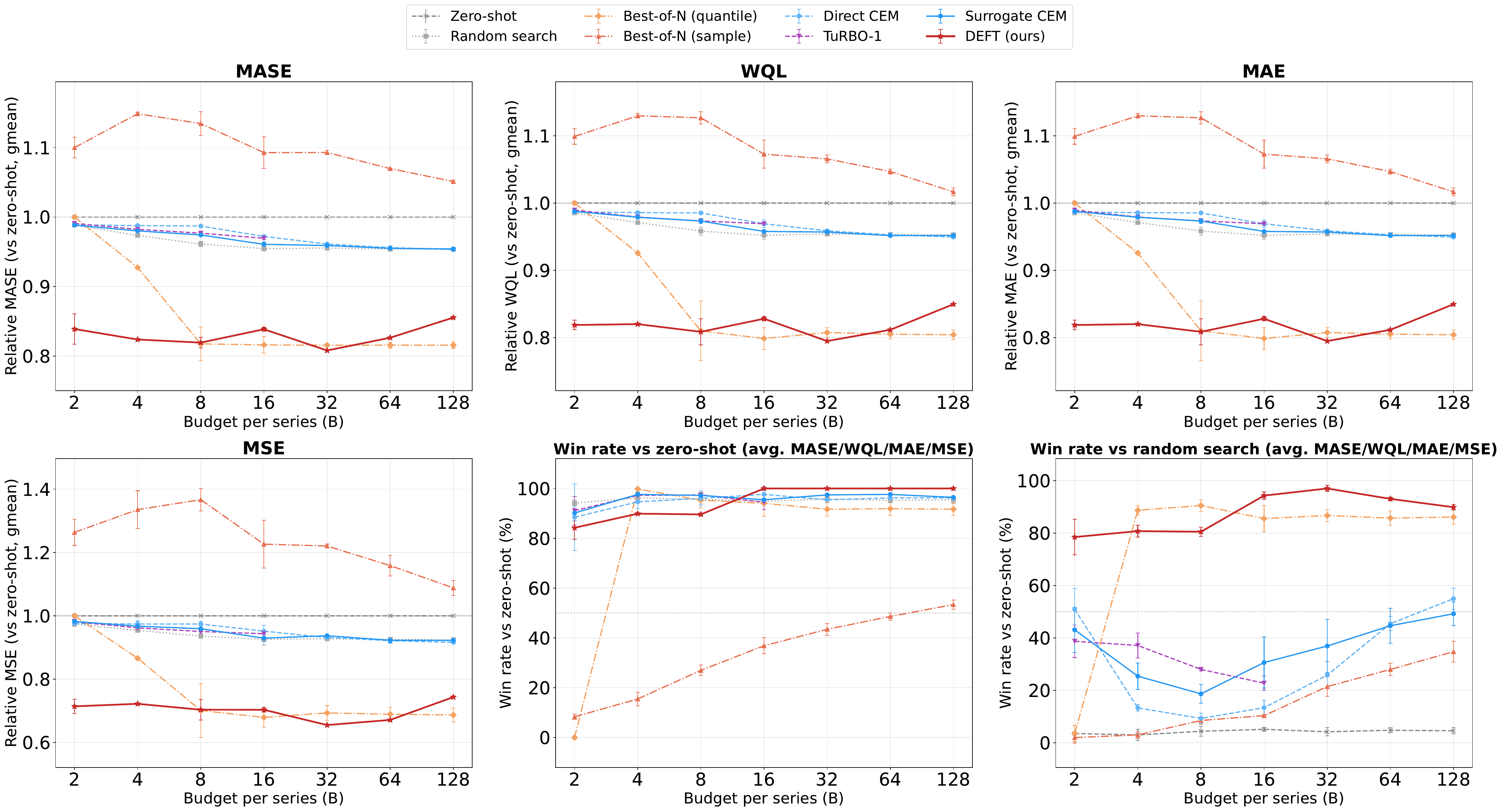}
    \caption{Performance across expert-query budgets on the Chronos Benchmark using the TimesFM foundation model with Pairwise expert feedback.}
    \label{fig:budget-chronosbm-timesfm-pairwise-noise0}
\end{figure*}

\begin{figure*}[h]
    \centering
    \includegraphics[width=\textwidth]{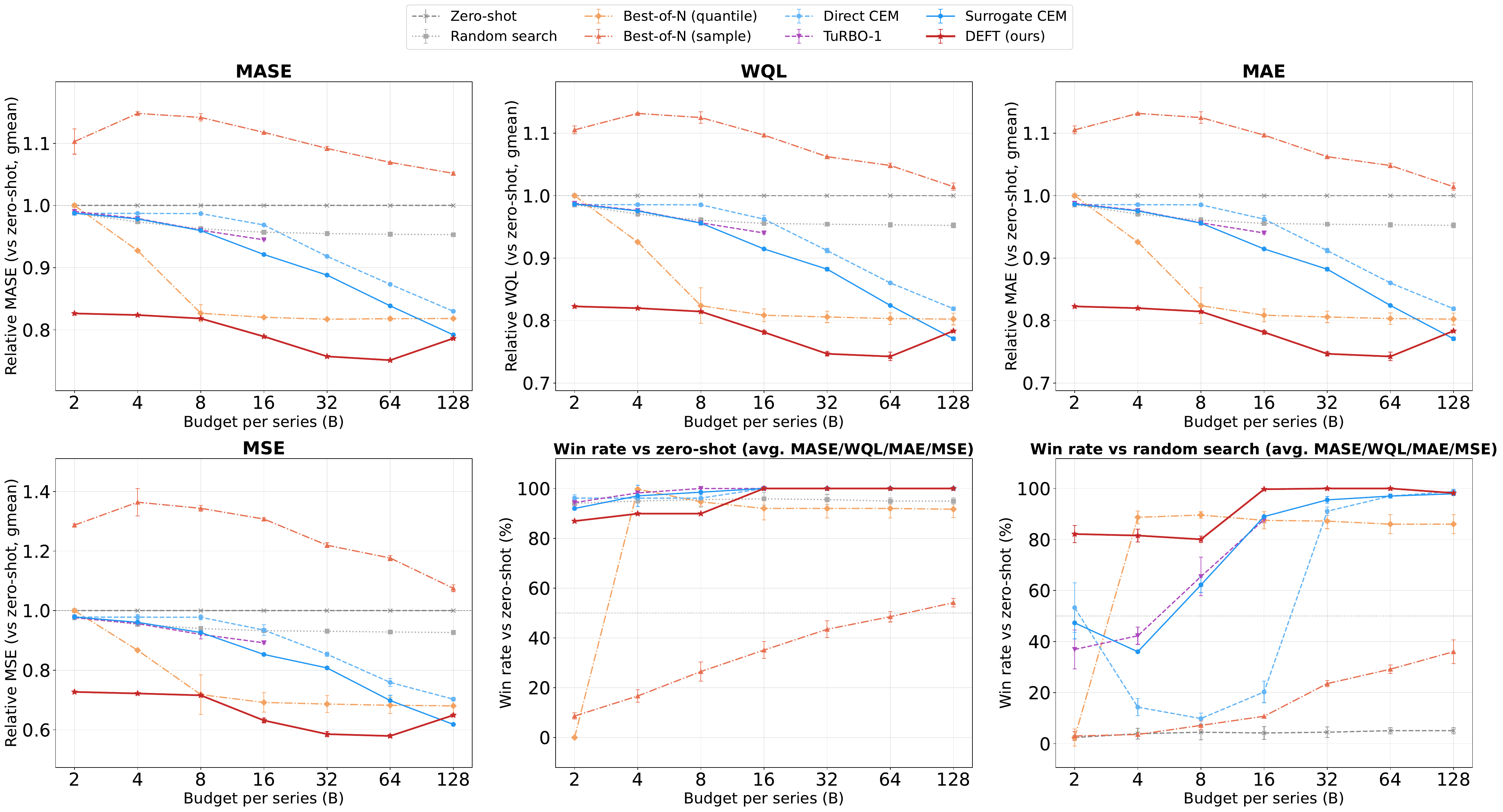}
    \caption{Performance across expert-query budgets on the Chronos Benchmark using the TimesFM foundation model with Pairwise-Best expert feedback.}
    \label{fig:budget-chronosbm-timesfm-pairwisebest-noise0}
\end{figure*}

\begin{figure*}[h]
    \centering
    \includegraphics[width=\textwidth]{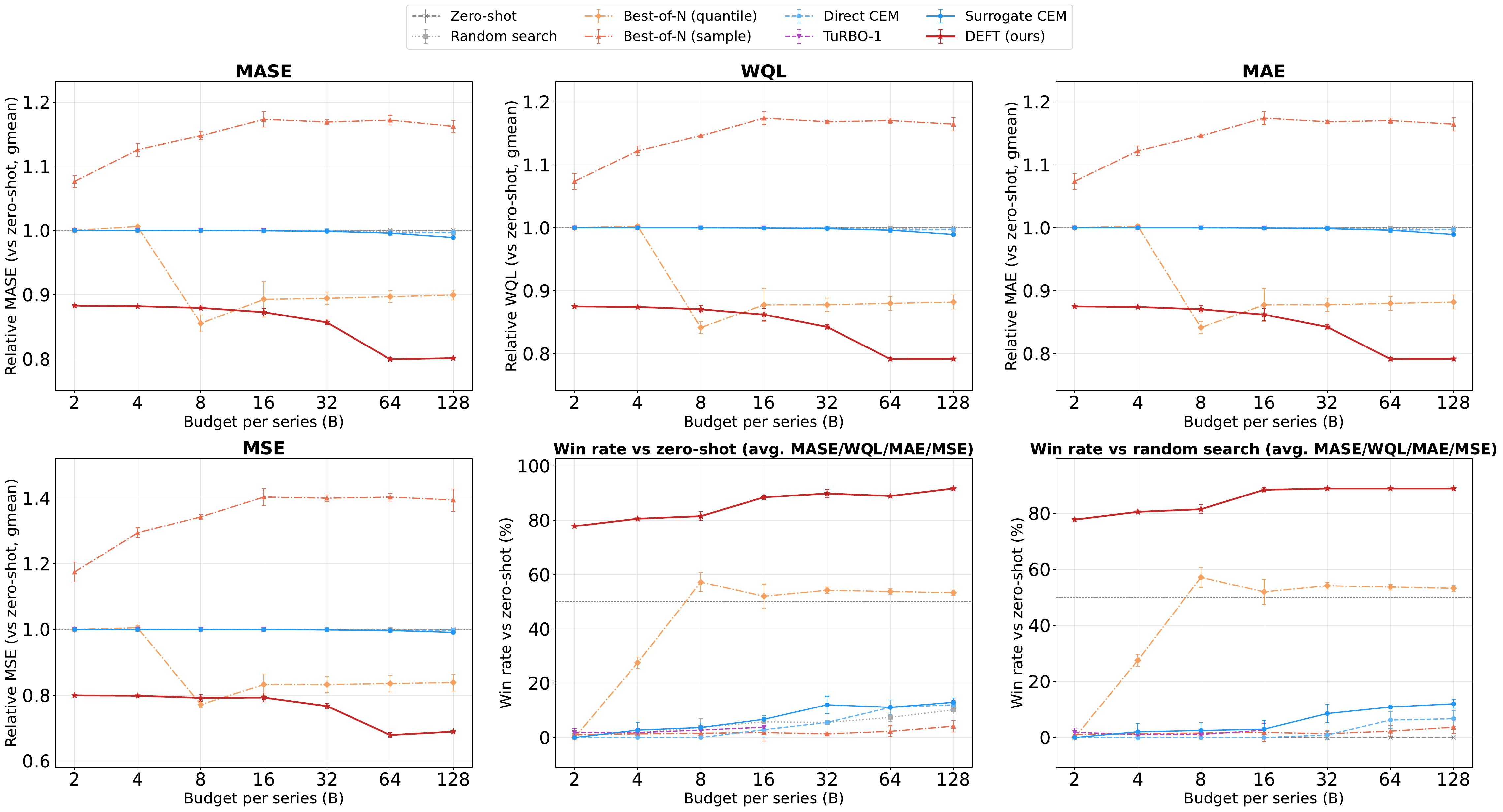}
    \caption{Performance across expert-query budgets on the GIFT-Eval Benchmark using the TimesFM foundation model with 3-Level Rating expert feedback.}
    \label{fig:budget-giftevalbm-timesfm-rating3-noise0}
\end{figure*}

\begin{figure*}[h]
    \centering
    \includegraphics[width=\textwidth]{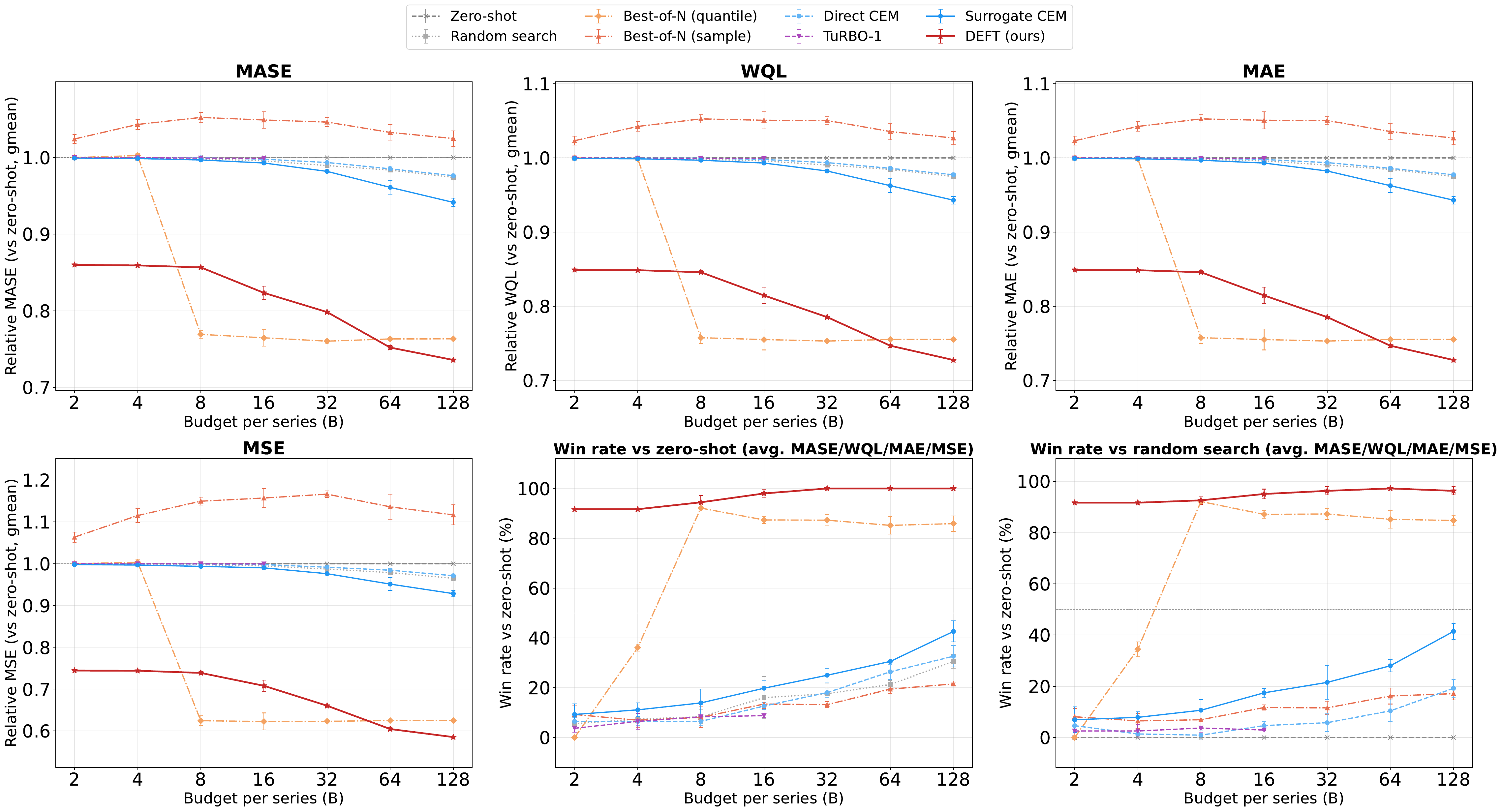}
    \caption{Performance across expert-query budgets on the GIFT-Eval Benchmark using the TimesFM foundation model with 5-Level Rating expert feedback.}
    \label{fig:budget-giftevalbm-timesfm-rating5-noise0}
\end{figure*}

\begin{figure*}[h]
    \centering
    \includegraphics[width=\textwidth]{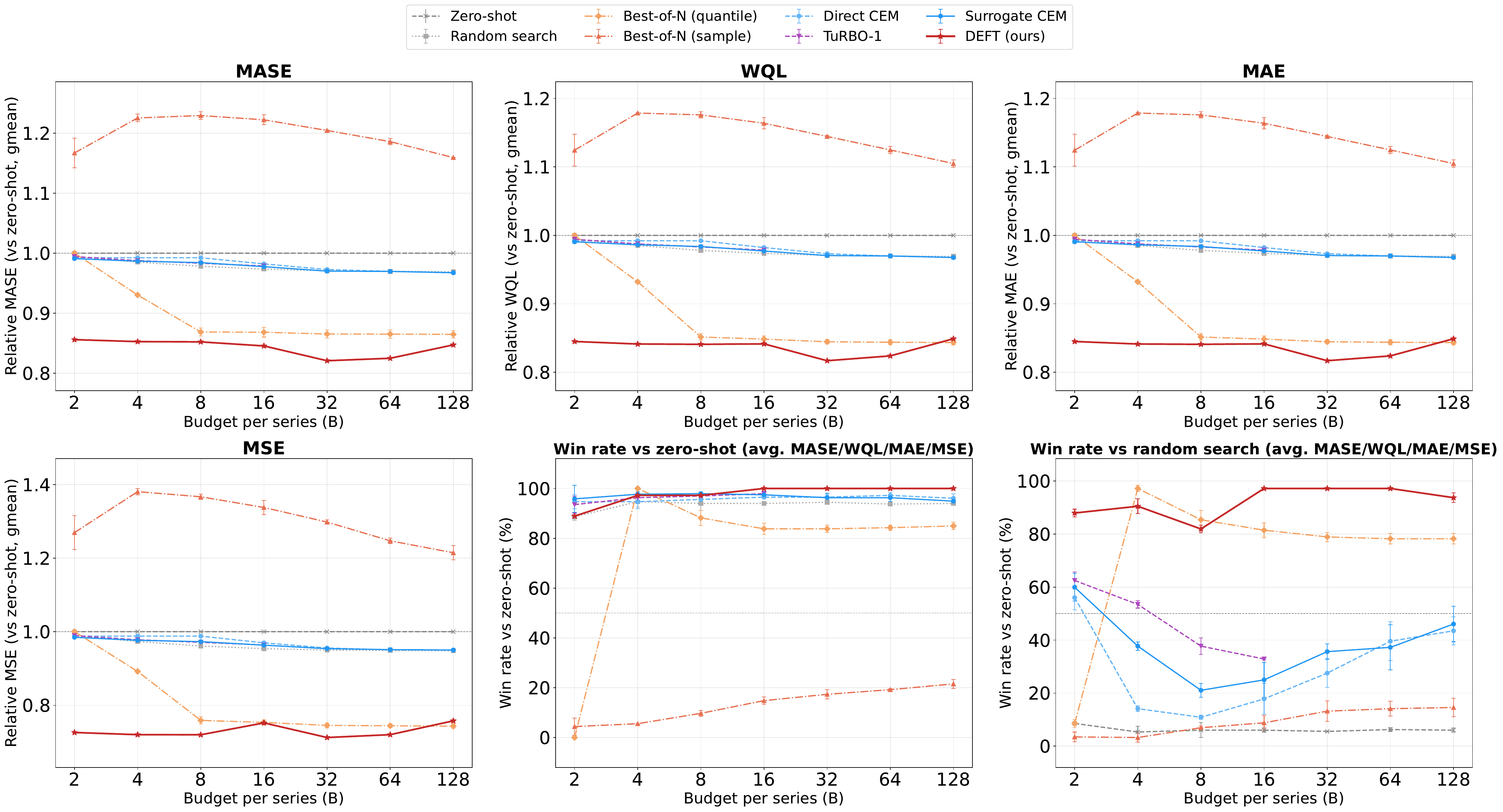}
    \caption{Performance across expert-query budgets on the GIFT-Eval Benchmark using the TimesFM foundation model with Pairwise expert feedback.}
    \label{fig:budget-giftevalbm-timesfm-pairwise-noise0}
\end{figure*}

\begin{figure*}[h]
    \centering
    \includegraphics[width=\textwidth]{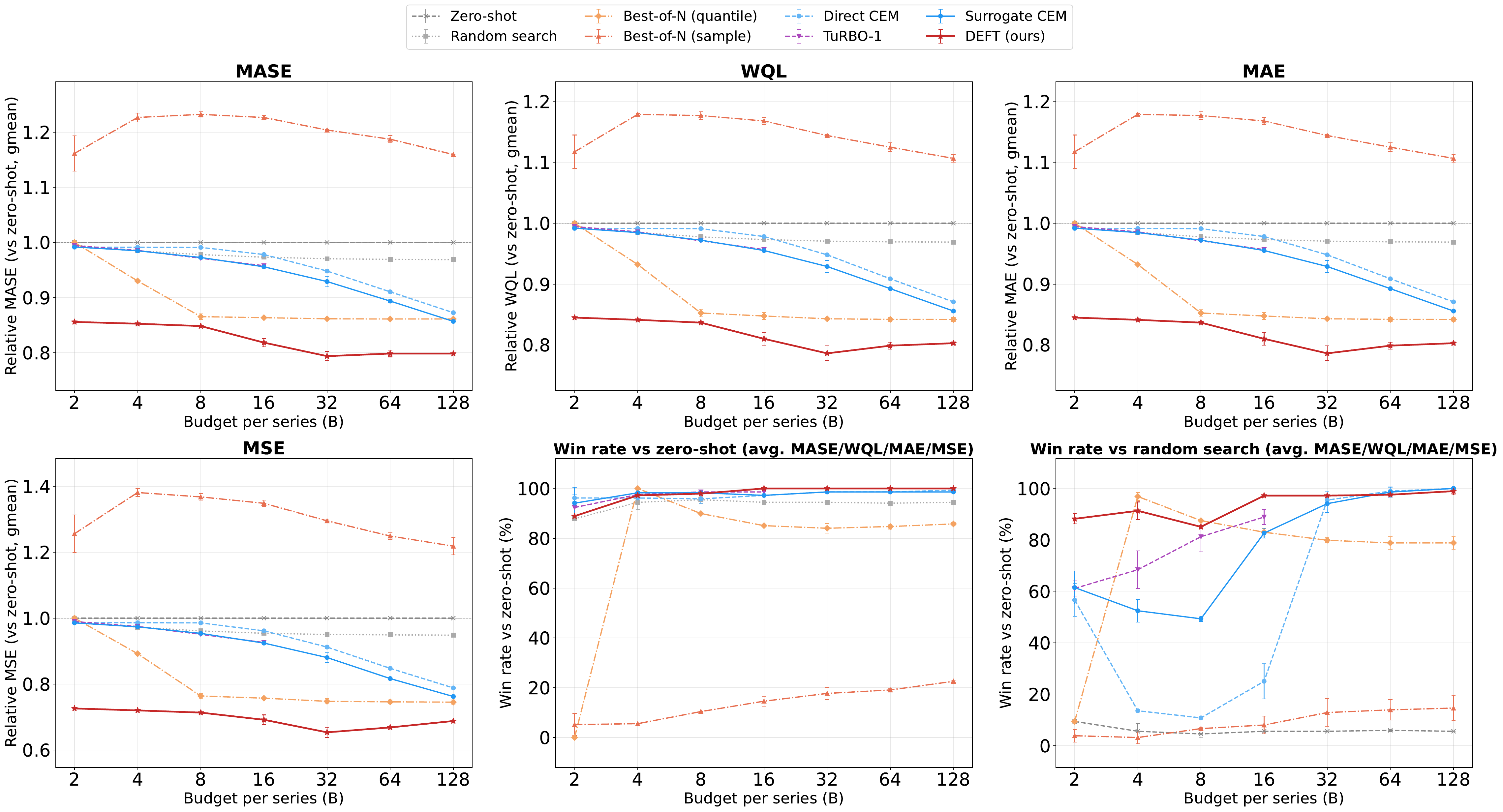}
    \caption{Performance across expert-query budgets on the GIFT-Eval Benchmark using the TimesFM foundation model with Pairwise-Best expert feedback.}
    \label{fig:budget-giftevalbm-timesfm-pairwisebest-noise0}
\end{figure*}

For the TimesFM foundation model, we report the performance curves across expert-query budgets for different expert-feedback settings as follows:

\begin{itemize}
    \item Figure~\ref{fig:budget-chronosbm-timesfm-rating3-noise0}: \textbf{Benchmark:} Chronos Benchmark; \textbf{Foundation model:} TimesFM; \textbf{Expert feedback:} 3-Level Rating.
    
    \item Figure~\ref{fig:budget-chronosbm-timesfm-rating5-noise0}: \textbf{Benchmark:} Chronos Benchmark; \textbf{Foundation model:} TimesFM; \textbf{Expert feedback:} 5-Level Rating.
    
    \item Figure~\ref{fig:budget-chronosbm-timesfm-pairwise-noise0}: \textbf{Benchmark:} Chronos Benchmark; \textbf{Foundation model:} TimesFM; \textbf{Expert feedback:} Pairwise.
    
    \item Figure~\ref{fig:budget-chronosbm-timesfm-pairwisebest-noise0}: \textbf{Benchmark:} Chronos Benchmark; \textbf{Foundation model:} TimesFM; \textbf{Expert feedback:} Pairwise-Best.


     \item Figure~\ref{fig:budget-giftevalbm-timesfm-rating3-noise0}: \textbf{Benchmark:} GIFT-Eval  Benchmark; \textbf{Foundation model:} TimesFM; \textbf{Expert feedback:} 3-Level Rating.
    
    \item Figure~\ref{fig:budget-giftevalbm-timesfm-rating5-noise0}: \textbf{Benchmark:} GIFT-Eval  Benchmark; \textbf{Foundation model:} TimesFM; \textbf{Expert feedback:} 5-Level Rating.
    
    \item Figure~\ref{fig:budget-giftevalbm-timesfm-pairwise-noise0}: \textbf{Benchmark:} GIFT-Eval  Benchmark; \textbf{Foundation model:} TimesFM; \textbf{Expert feedback:} Pairwise.
    
    \item Figure~\ref{fig:budget-giftevalbm-timesfm-pairwisebest-noise0}: \textbf{Benchmark:} GIFT-Eval Benchmark; \textbf{Foundation model:} TimesFM; \textbf{Expert feedback:} Pairwise-Best.

    
\end{itemize}

\begin{figure*}[h]
    \centering
    \includegraphics[width=\textwidth]{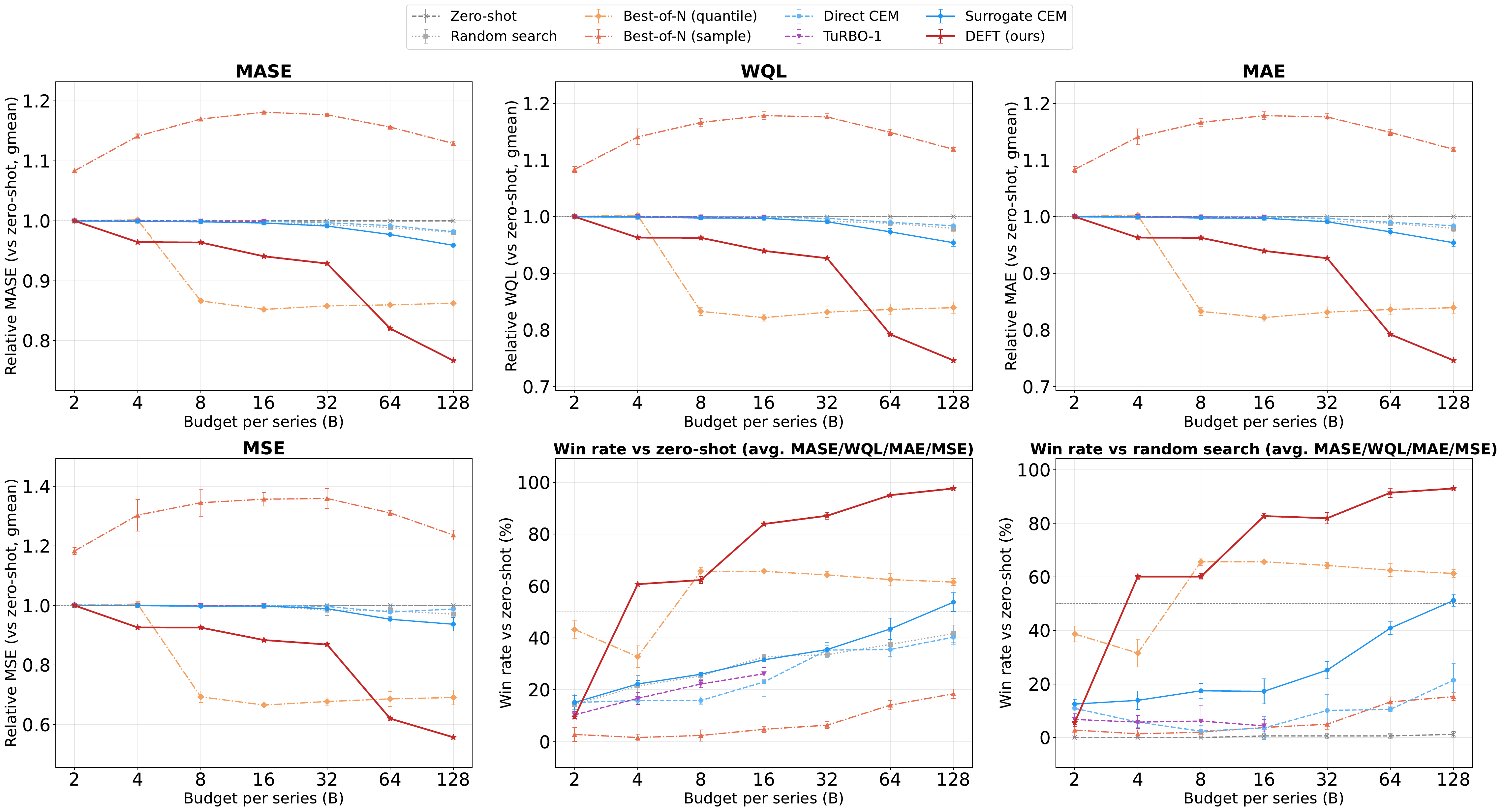}
    \caption{Performance across expert-query budgets on the Chronos Benchmark using the Chronos foundation model with 3-Level Rating expert feedback.}
    \label{fig:budget-chronosbm-chronos-rating3-noise0}
\end{figure*}

\begin{figure*}[h]
    \centering
    \includegraphics[width=\textwidth]{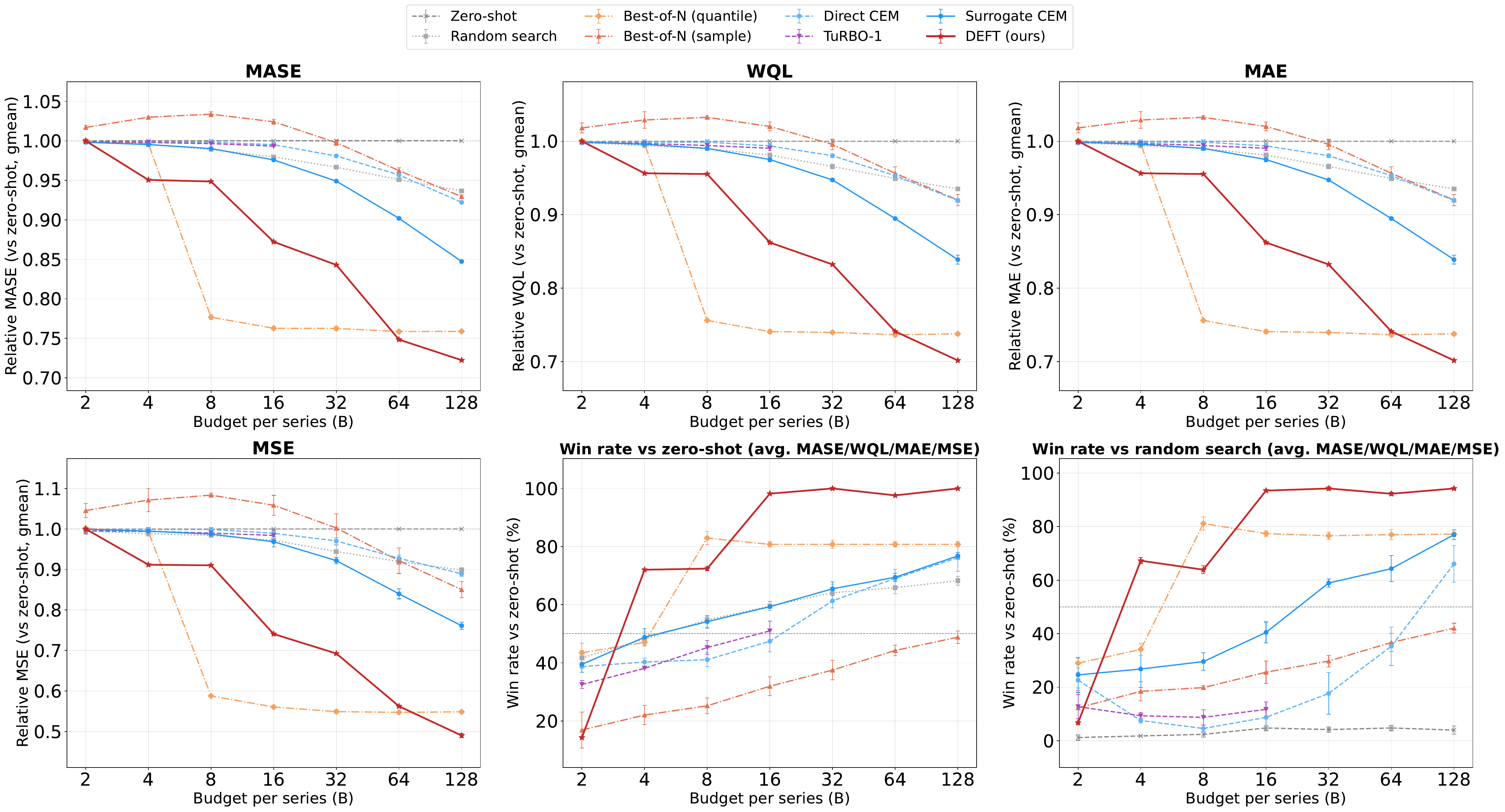}
    \caption{Performance across expert-query budgets on the Chronos Benchmark using the Chronos foundation model with 5-Level Rating expert feedback.}
    \label{fig:budget-chronosbm-chronos-rating5-noise0}
\end{figure*}

\begin{figure*}[h]
    \centering
    \includegraphics[width=\textwidth]{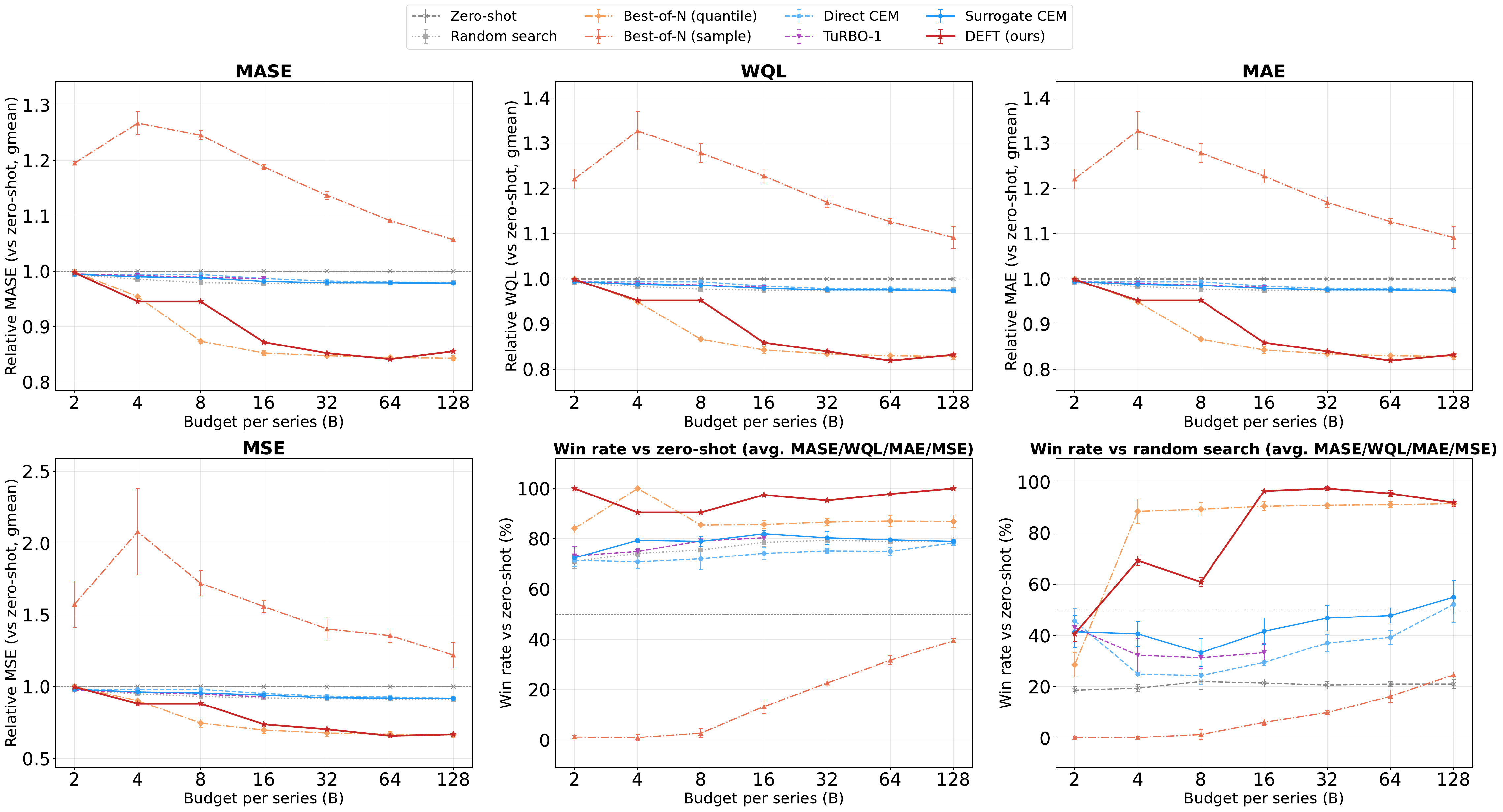}
    \caption{Performance across expert-query budgets on the Chronos Benchmark using the Chronos foundation model with Pairwise expert feedback.}
    \label{fig:budget-chronosbm-chronos-pairwise-noise0}
\end{figure*}

\begin{figure*}[h]
    \centering
    \includegraphics[width=\textwidth]{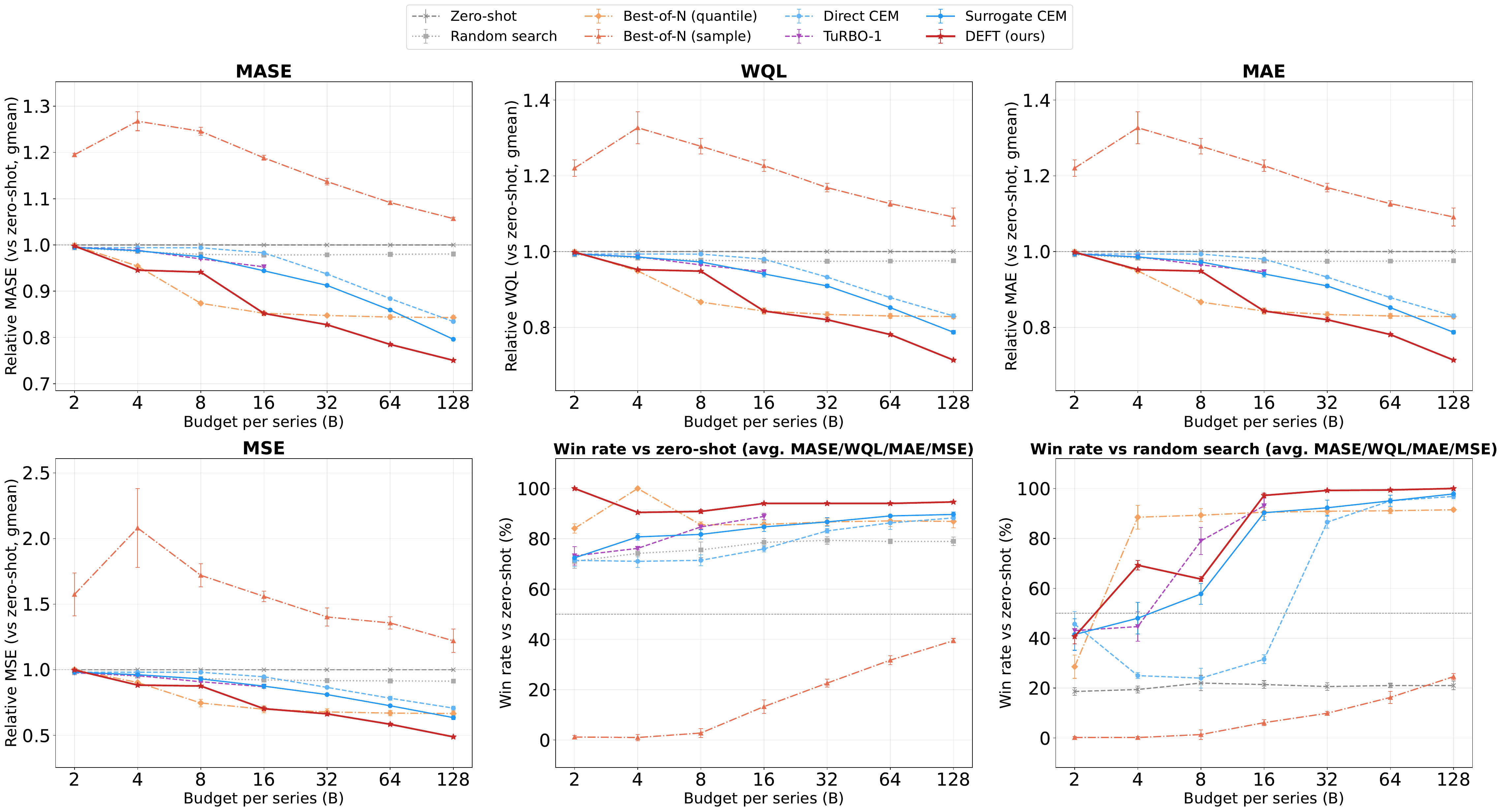}
    \caption{Performance across expert-query budgets on the Chronos Benchmark using the Chronos foundation model with Pairwise-Best expert feedback.}
    \label{fig:budget-chronosbm-chronos-pairwisebest-noise0}
\end{figure*}

\begin{figure*}[h]
    \centering
    \includegraphics[width=\textwidth]{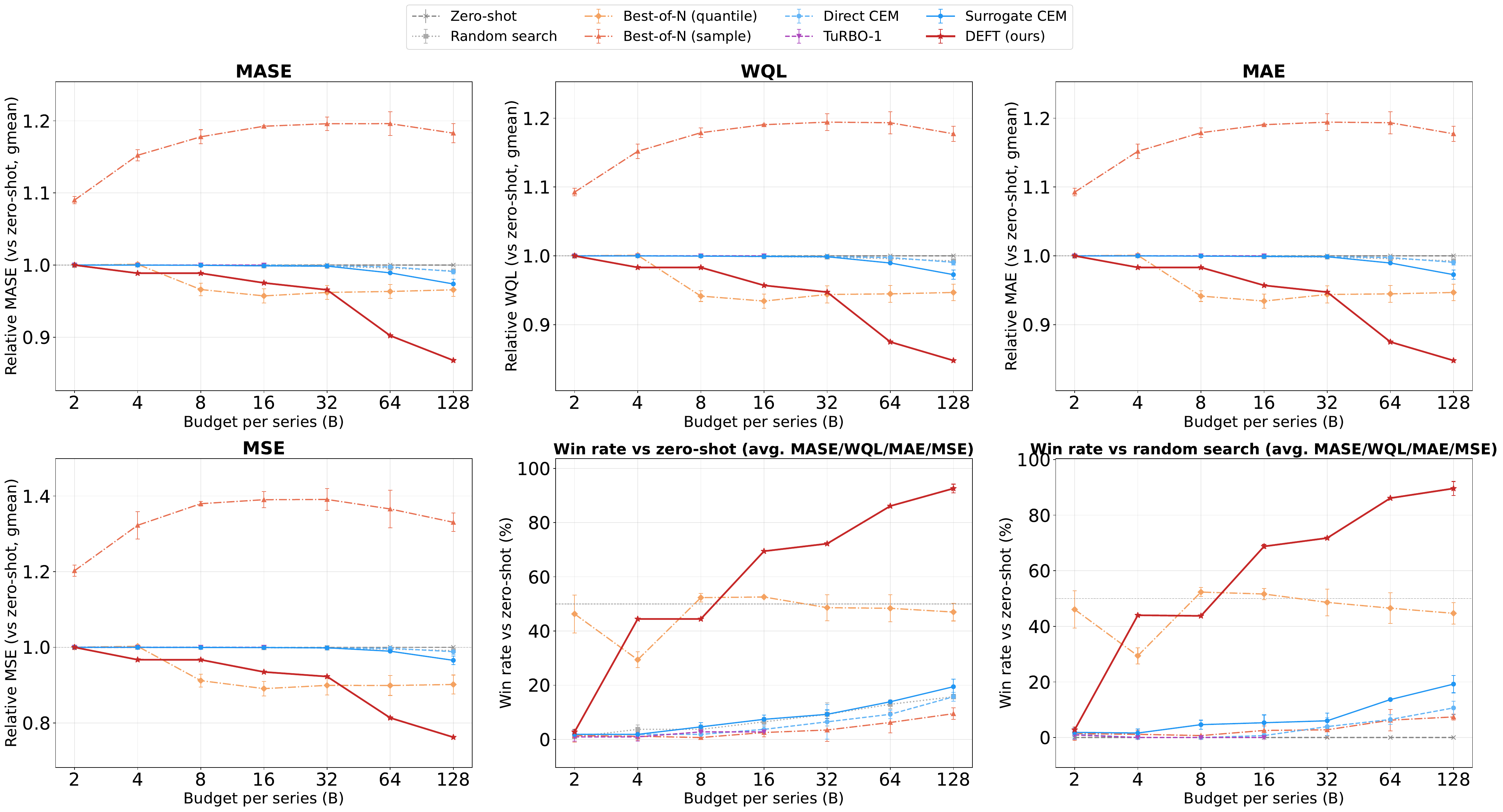}
    \caption{Performance across expert-query budgets on the GIFT-Eval Benchmark using the Chronos foundation model with 3-Level Rating expert feedback.}
    \label{fig:budget-giftevalbm-chronos-rating3-noise0}
\end{figure*}

\begin{figure*}[h]
    \centering
    \includegraphics[width=\textwidth]{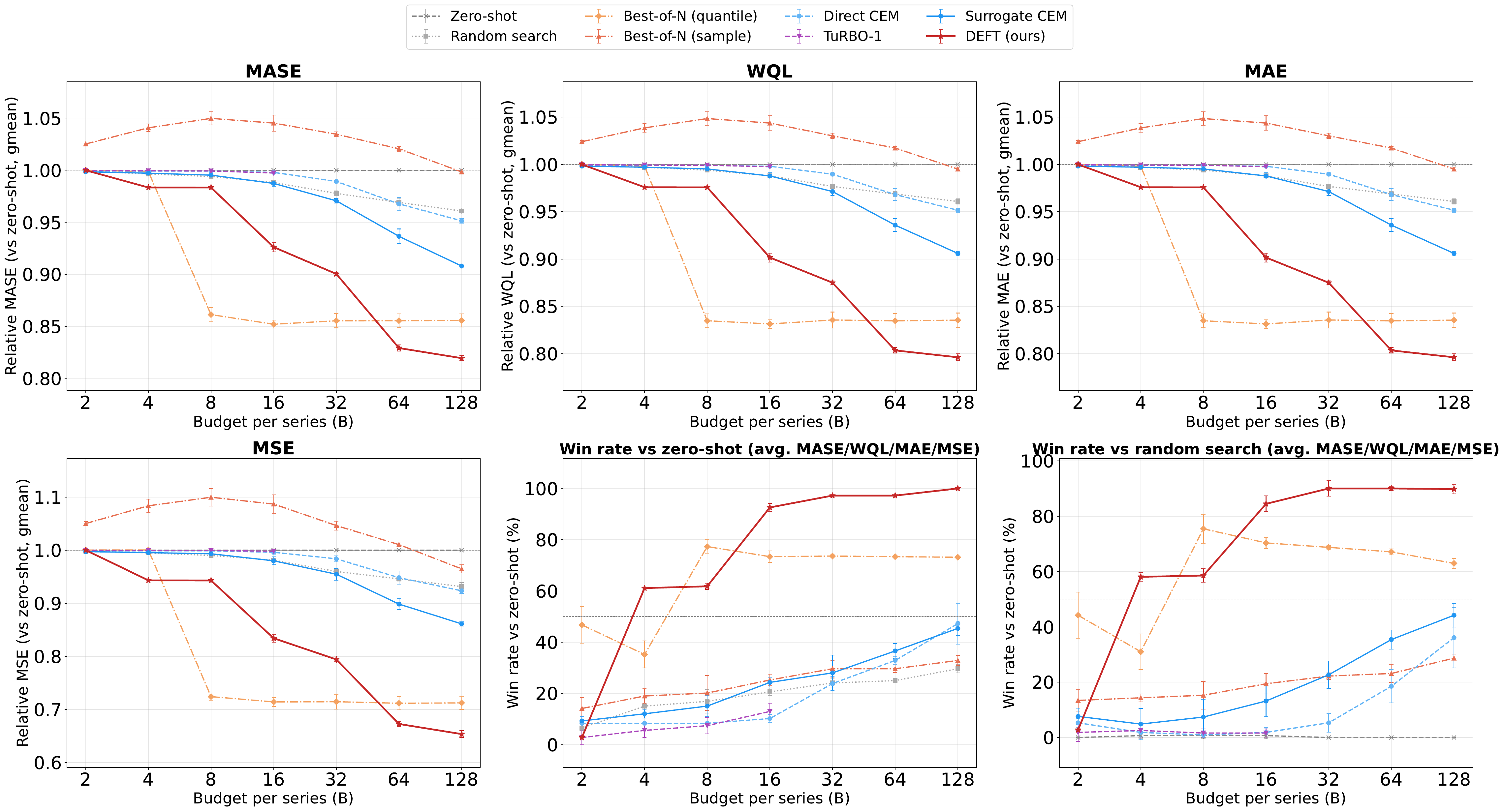}
    \caption{Performance across expert-query budgets on the GIFT-Eval Benchmark using the Chronos foundation model with 5-Level Rating expert feedback.}
    \label{fig:budget-giftevalbm-chronos-rating5-noise0}
\end{figure*}

\begin{figure*}[h]
    \centering
    \includegraphics[width=\textwidth]{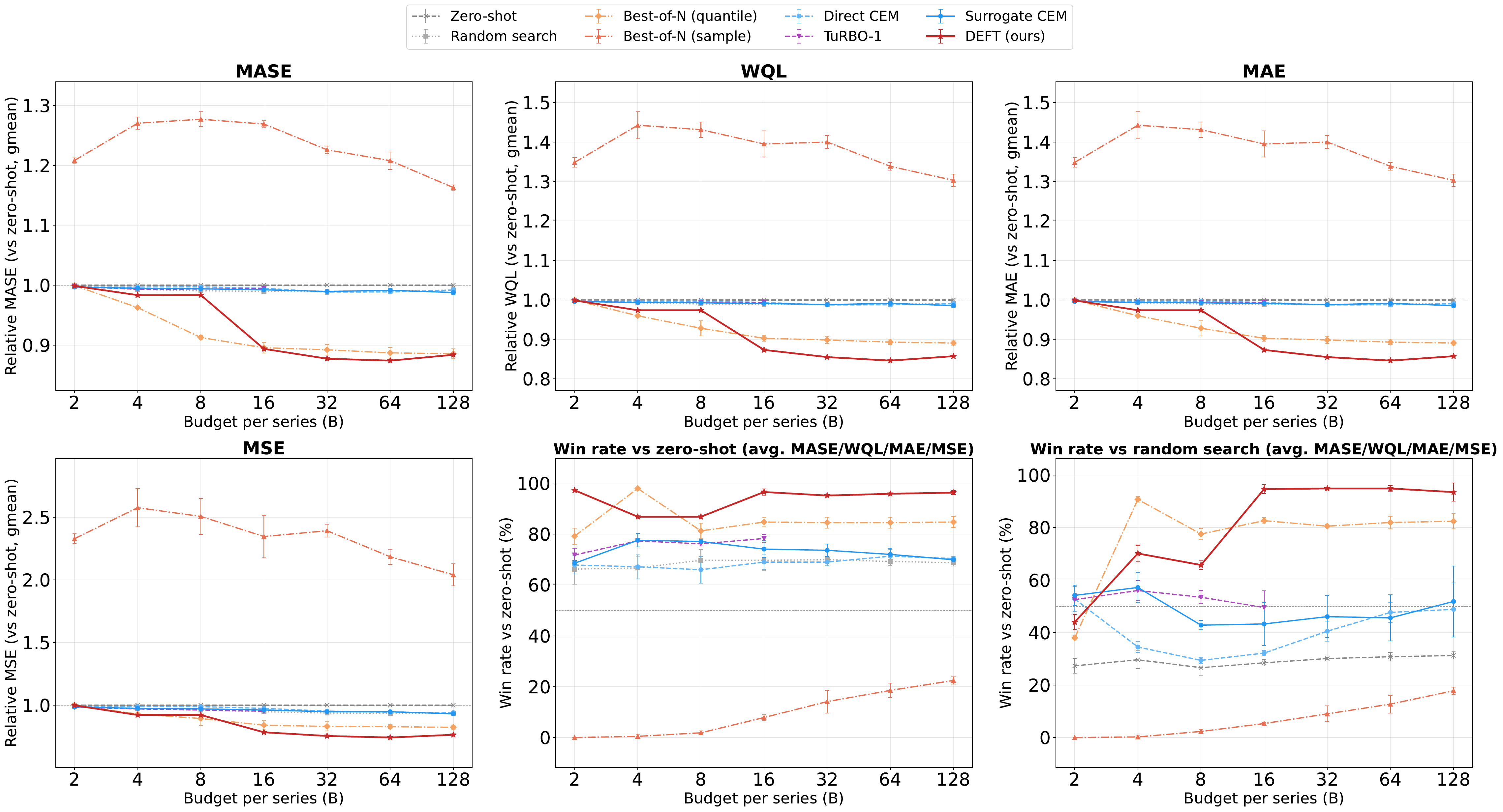}
    \caption{Performance across expert-query budgets on the GIFT-Eval Benchmark using the Chronos foundation model with Pairwise expert feedback.}
    \label{fig:budget-giftevalbm-chronos-pairwise-noise0}
\end{figure*}

\begin{figure*}[h]
    \centering
    \includegraphics[width=\textwidth]{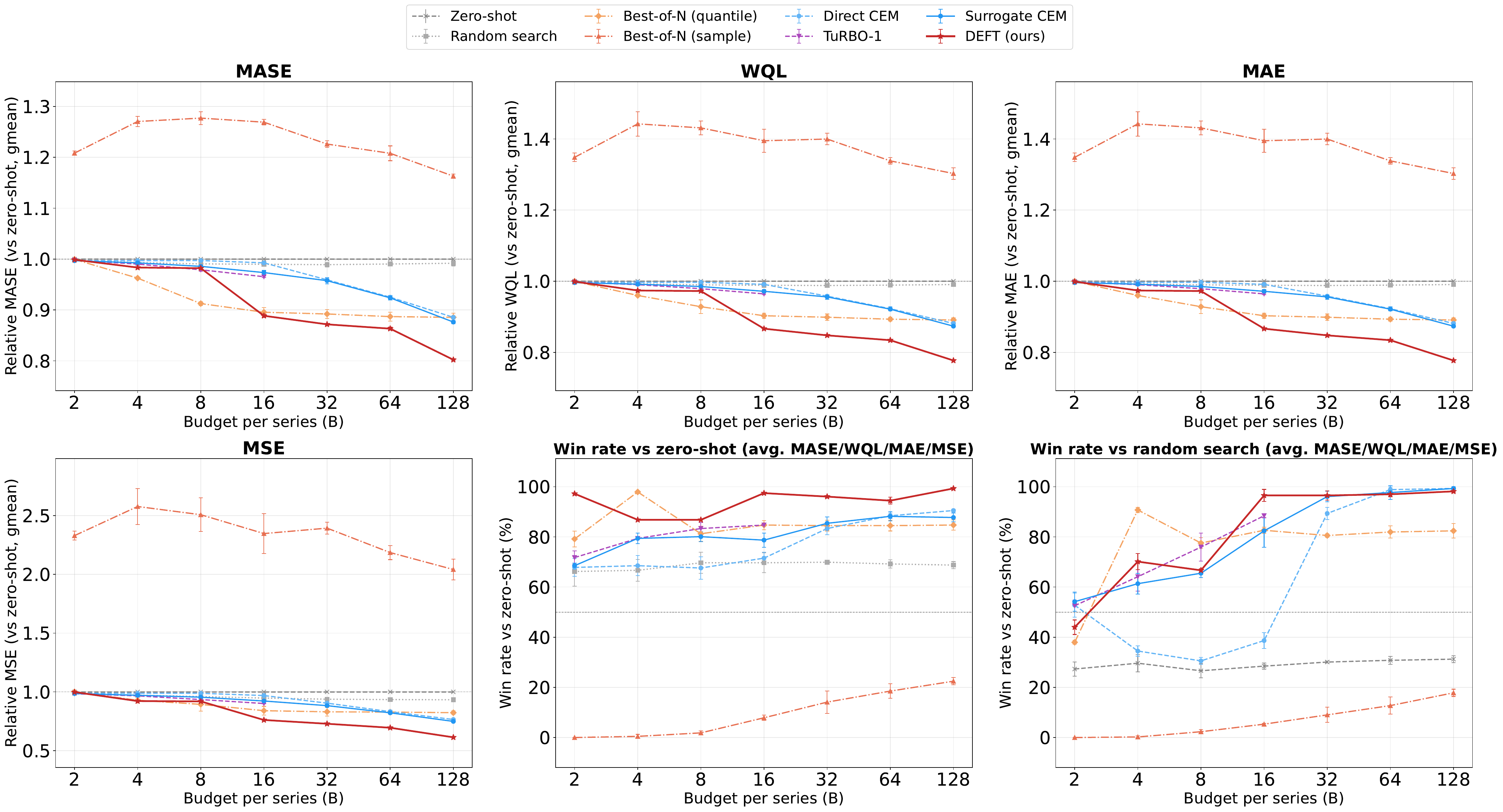}
    \caption{Performance across expert-query budgets on the GIFT-Eval Benchmark using the Chronos foundation model with Pairwise-Best expert feedback.}
    \label{fig:budget-giftevalbm-chronos-pairwisebest-noise0}
\end{figure*}

For the Chronos foundation model, we report the performance curves across expert-query budgets for different expert-feedback settings as follows:

\begin{itemize}
    \item Figure~\ref{fig:budget-chronosbm-chronos-rating3-noise0}: \textbf{Benchmark:} Chronos Benchmark; \textbf{Foundation model:} Chronos; \textbf{Expert feedback:} 3-Level Rating.

    \item Figure~\ref{fig:budget-chronosbm-chronos-rating5-noise0}: \textbf{Benchmark:} Chronos Benchmark; \textbf{Foundation model:} Chronos; \textbf{Expert feedback:} 5-Level Rating.

    \item Figure~\ref{fig:budget-chronosbm-chronos-pairwise-noise0}: \textbf{Benchmark:} Chronos Benchmark; \textbf{Foundation model:} Chronos; \textbf{Expert feedback:} Pairwise.

    \item Figure~\ref{fig:budget-chronosbm-chronos-pairwisebest-noise0}: \textbf{Benchmark:} Chronos Benchmark; \textbf{Foundation model:} Chronos; \textbf{Expert feedback:} Pairwise-Best.


    \item Figure~\ref{fig:budget-giftevalbm-chronos-rating3-noise0}: \textbf{Benchmark:} GIFT-Eval Benchmark; \textbf{Foundation model:} Chronos; \textbf{Expert feedback:} 3-Level Rating.

    \item Figure~\ref{fig:budget-giftevalbm-chronos-rating5-noise0}: \textbf{Benchmark:} GIFT-Eval Benchmark; \textbf{Foundation model:} Chronos; \textbf{Expert feedback:} 5-Level Rating.

    \item Figure~\ref{fig:budget-giftevalbm-chronos-pairwise-noise0}: \textbf{Benchmark:} GIFT-Eval Benchmark; \textbf{Foundation model:} Chronos; \textbf{Expert feedback:} Pairwise.

    \item Figure~\ref{fig:budget-giftevalbm-chronos-pairwisebest-noise0}: \textbf{Benchmark:} GIFT-Eval Benchmark; \textbf{Foundation model:} Chronos; \textbf{Expert feedback:} Pairwise-Best.

\end{itemize}

\begin{figure*}[h]
    \centering
    \includegraphics[width=\textwidth]{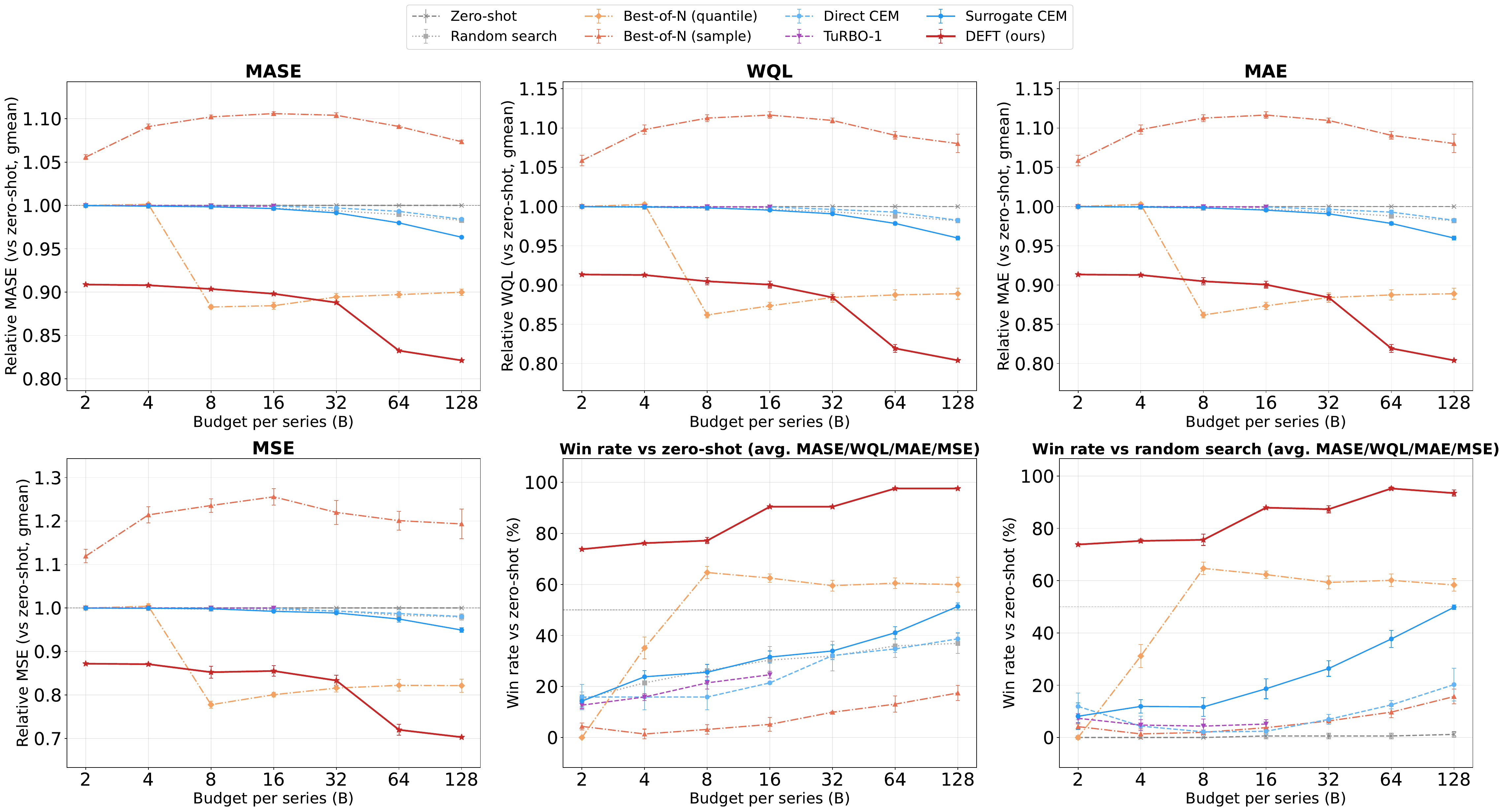}
    \caption{Performance across expert-query budgets on the Chronos Benchmark using the Moirai foundation model with 3-Level Rating expert feedback.}
    \label{fig:budget-chronosbm-moirai-rating3-noise0}
\end{figure*}

\begin{figure*}[h]
    \centering
    \includegraphics[width=\textwidth]{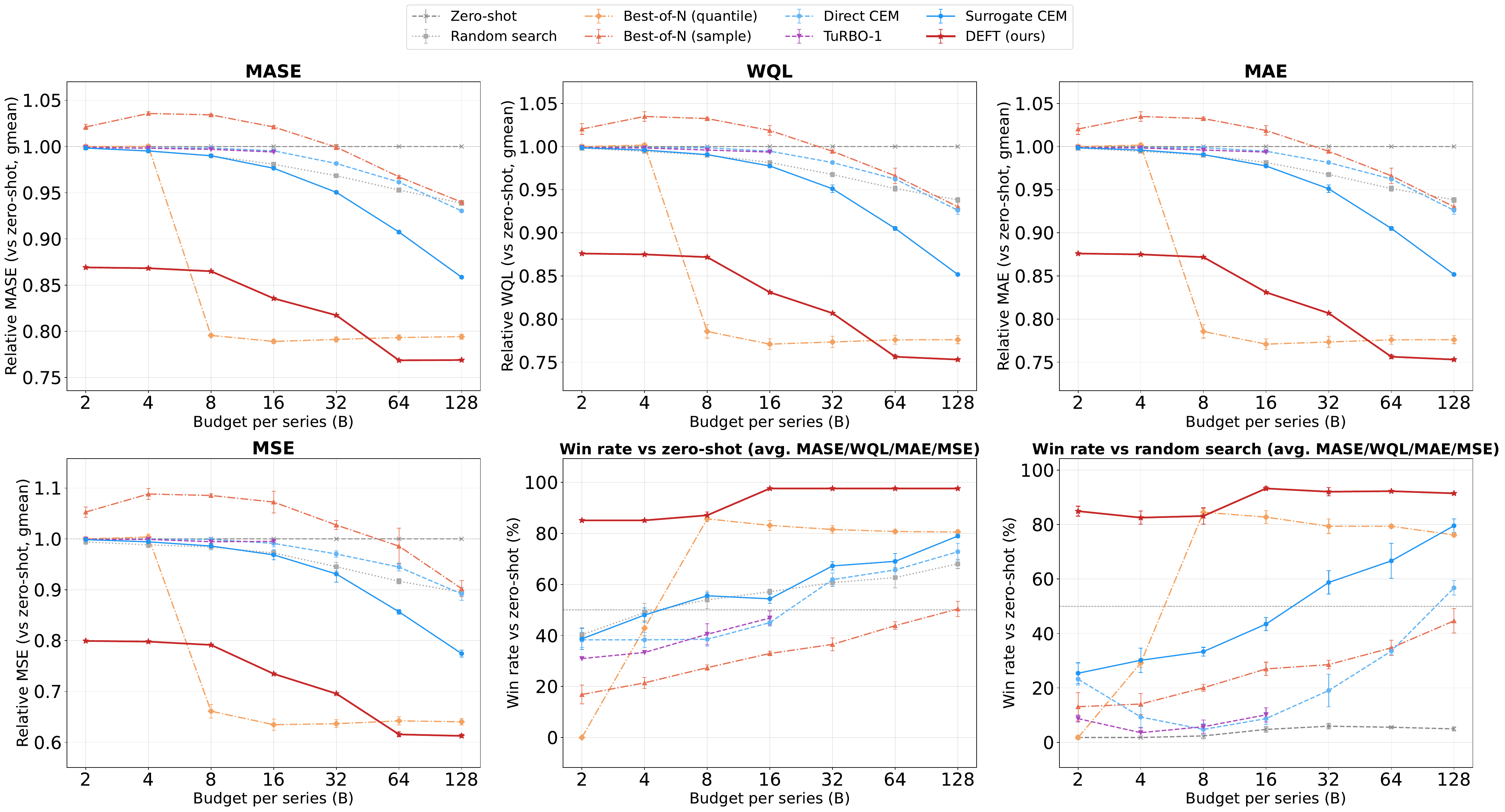}
    \caption{Performance across expert-query budgets on the Chronos Benchmark using the Moirai foundation model with 5-Level Rating expert feedback.}
    \label{fig:budget-chronosbm-moirai-rating5-noise0}
\end{figure*}

\begin{figure*}[h]
    \centering
    \includegraphics[width=\textwidth]{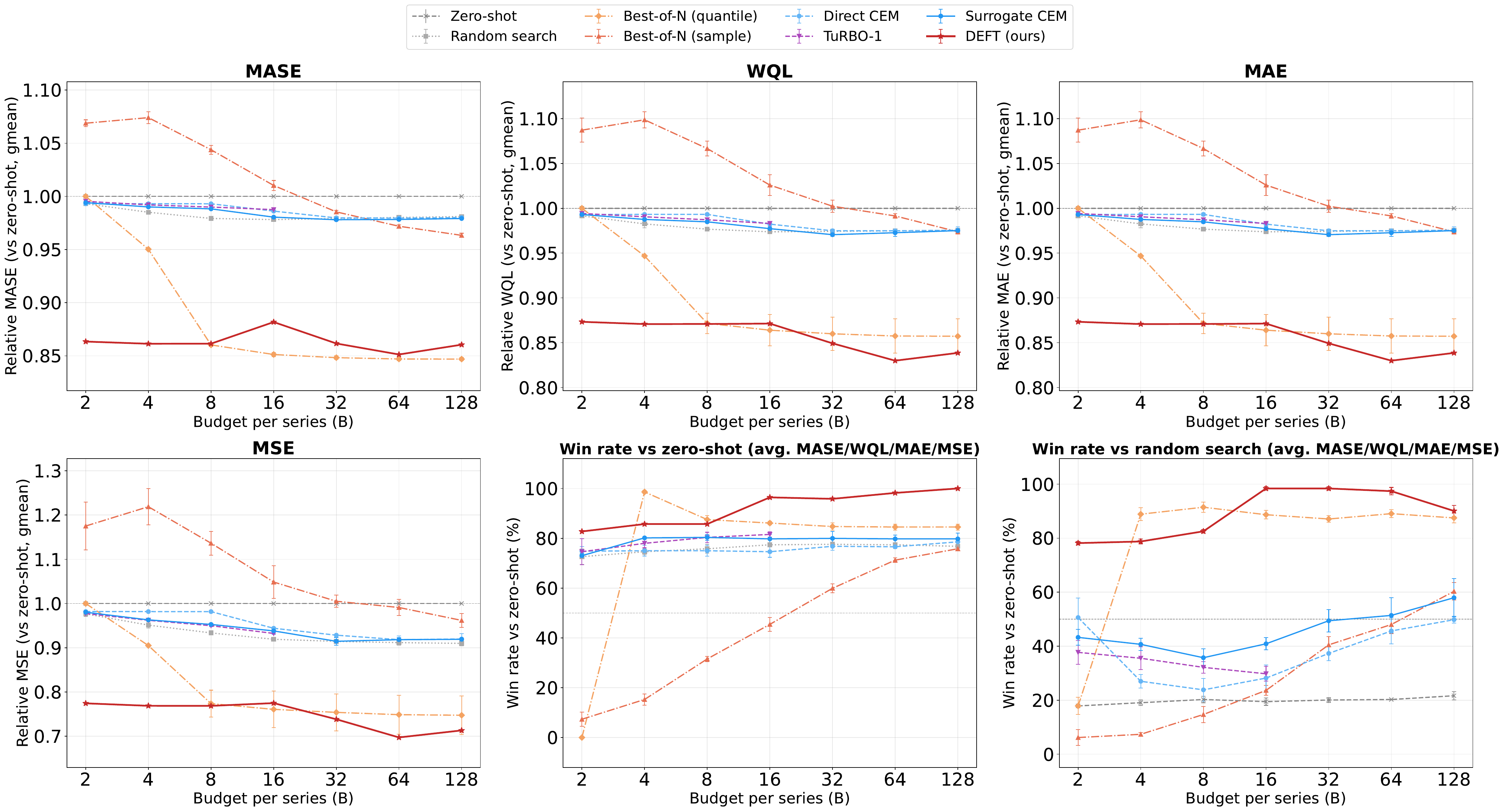}
    \caption{Performance across expert-query budgets on the Chronos Benchmark using the Moirai foundation model with Pairwise expert feedback.}
    \label{fig:budget-chronosbm-moirai-pairwise-noise0}
\end{figure*}

\begin{figure*}[h]
    \centering
    \includegraphics[width=\textwidth]{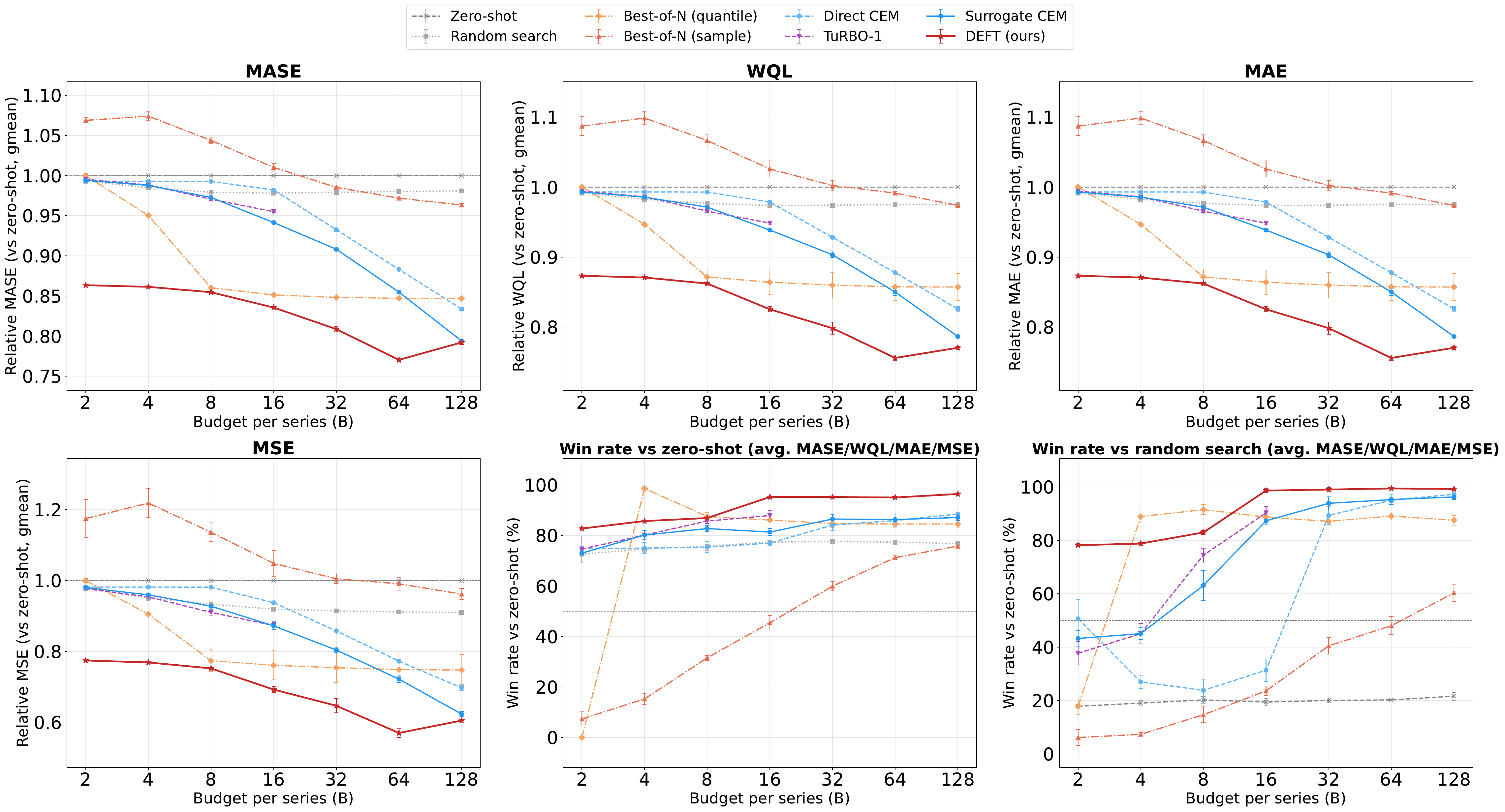}
    \caption{Performance across expert-query budgets on the Chronos Benchmark using the Moirai foundation model with Pairwise-Best expert feedback.}
    \label{fig:budget-chronosbm-moirai-pairwisebest-noise0}
\end{figure*}

\begin{figure*}[h]
    \centering
    \includegraphics[width=\textwidth]{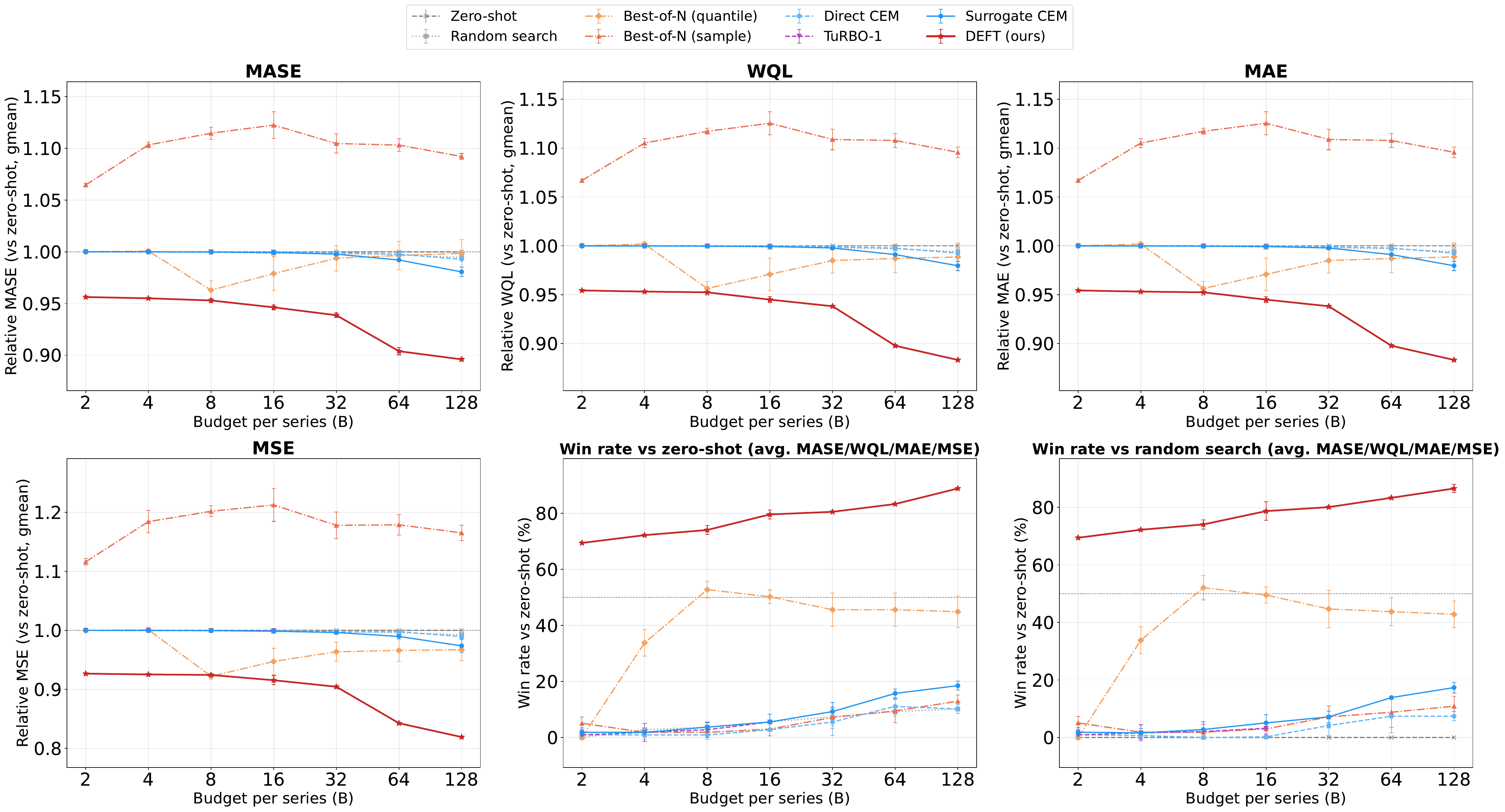}
    \caption{Performance across expert-query budgets on the GIFT-Eval Benchmark using the Moirai foundation model with 3-Level Rating expert feedback.}
    \label{fig:budget-giftevalbm-moirai-rating3-noise0}
\end{figure*}

\begin{figure*}[h]
    \centering
    \includegraphics[width=\textwidth]{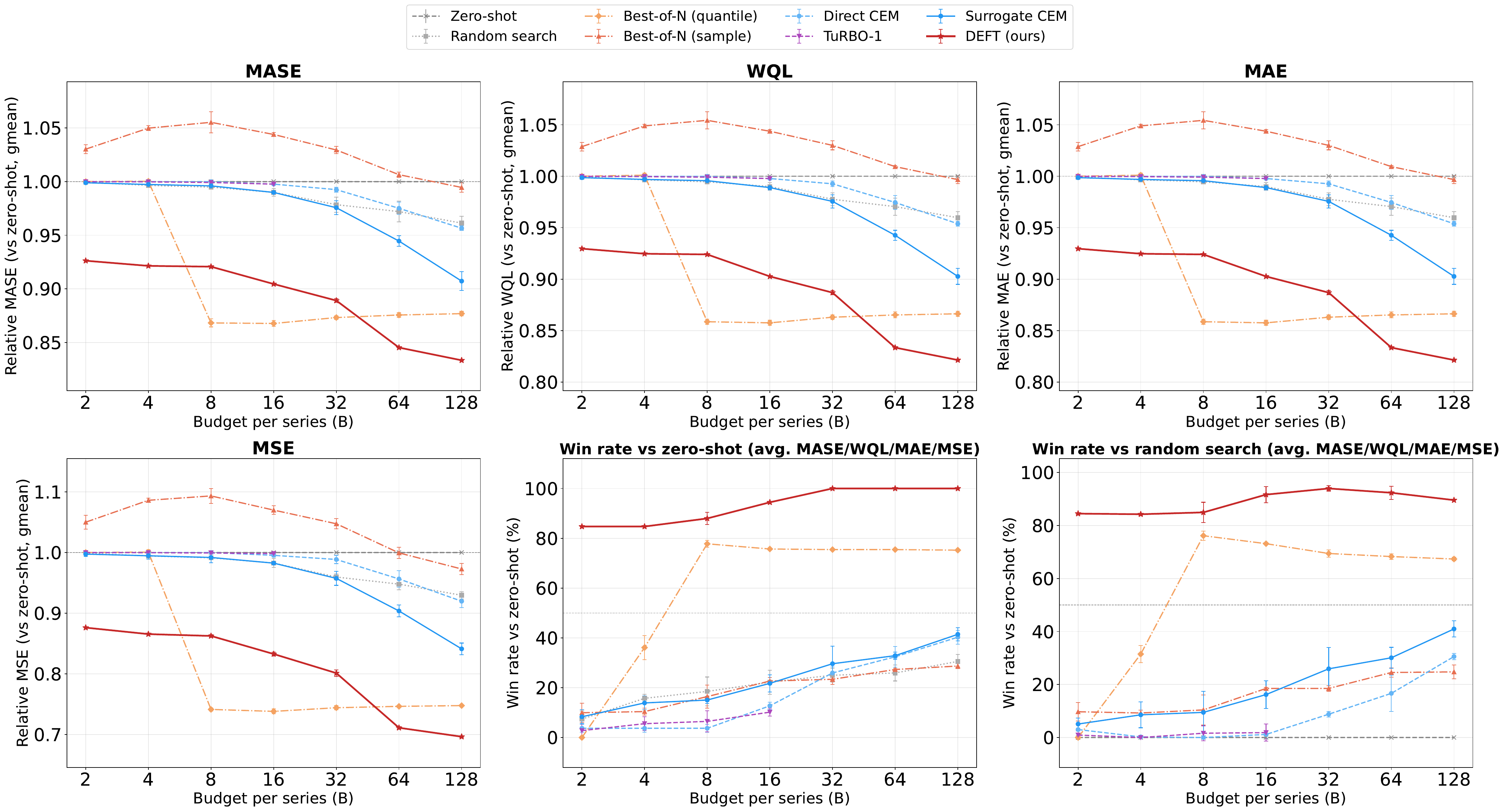}
    \caption{Performance across expert-query budgets on the GIFT-Eval Benchmark using the Moirai foundation model with 5-Level Rating expert feedback.}
    \label{fig:budget-giftevalbm-moirai-rating5-noise0}
\end{figure*}

\begin{figure*}[h]
    \centering
    \includegraphics[width=\textwidth]{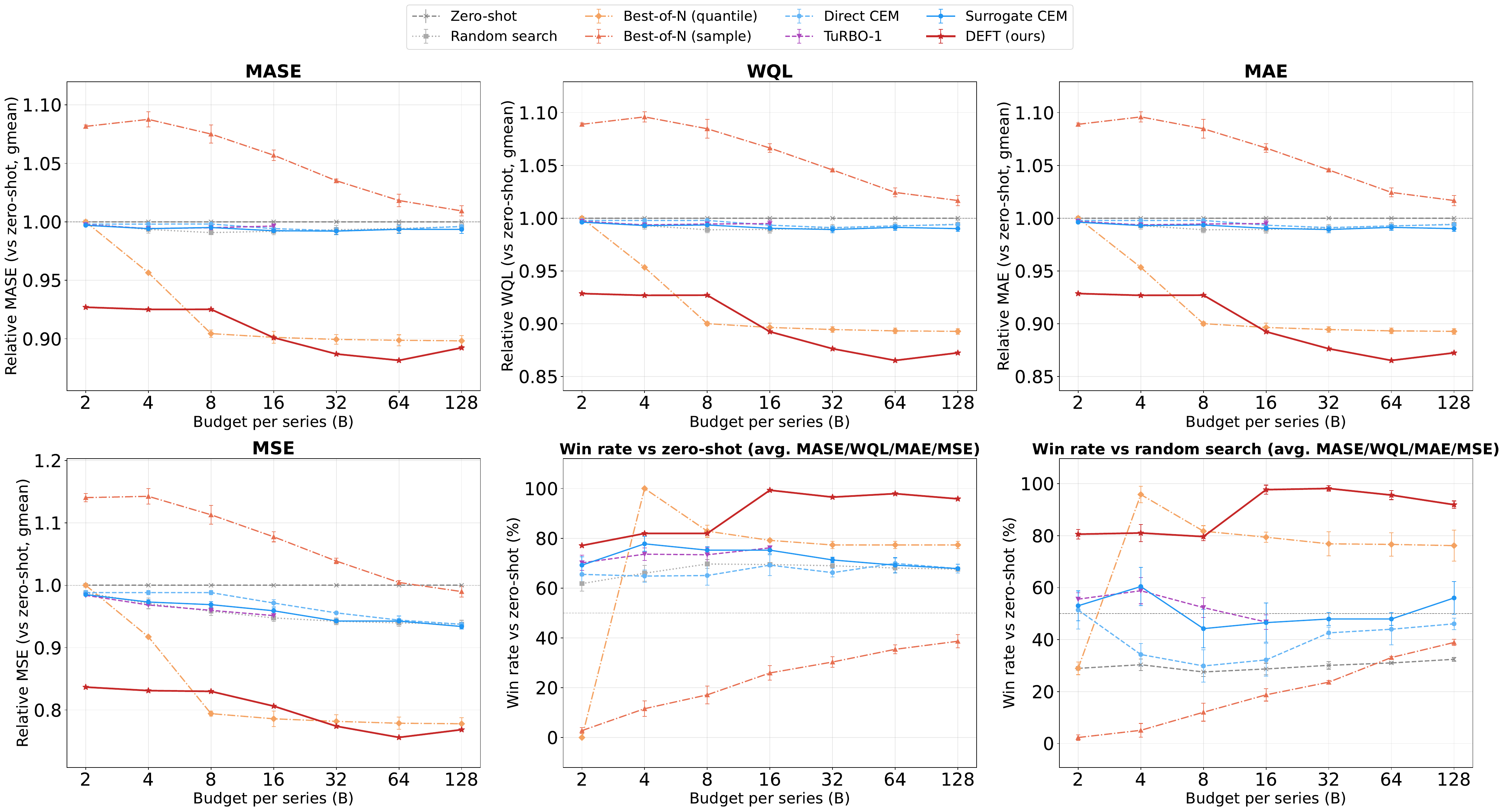}
    \caption{Performance across expert-query budgets on the GIFT-Eval Benchmark using the Moirai foundation model with Pairwise expert feedback.}
    \label{fig:budget-giftevalbm-moirai-pairwise-noise0}
\end{figure*}

\begin{figure*}[h]
    \centering
    \includegraphics[width=\textwidth]{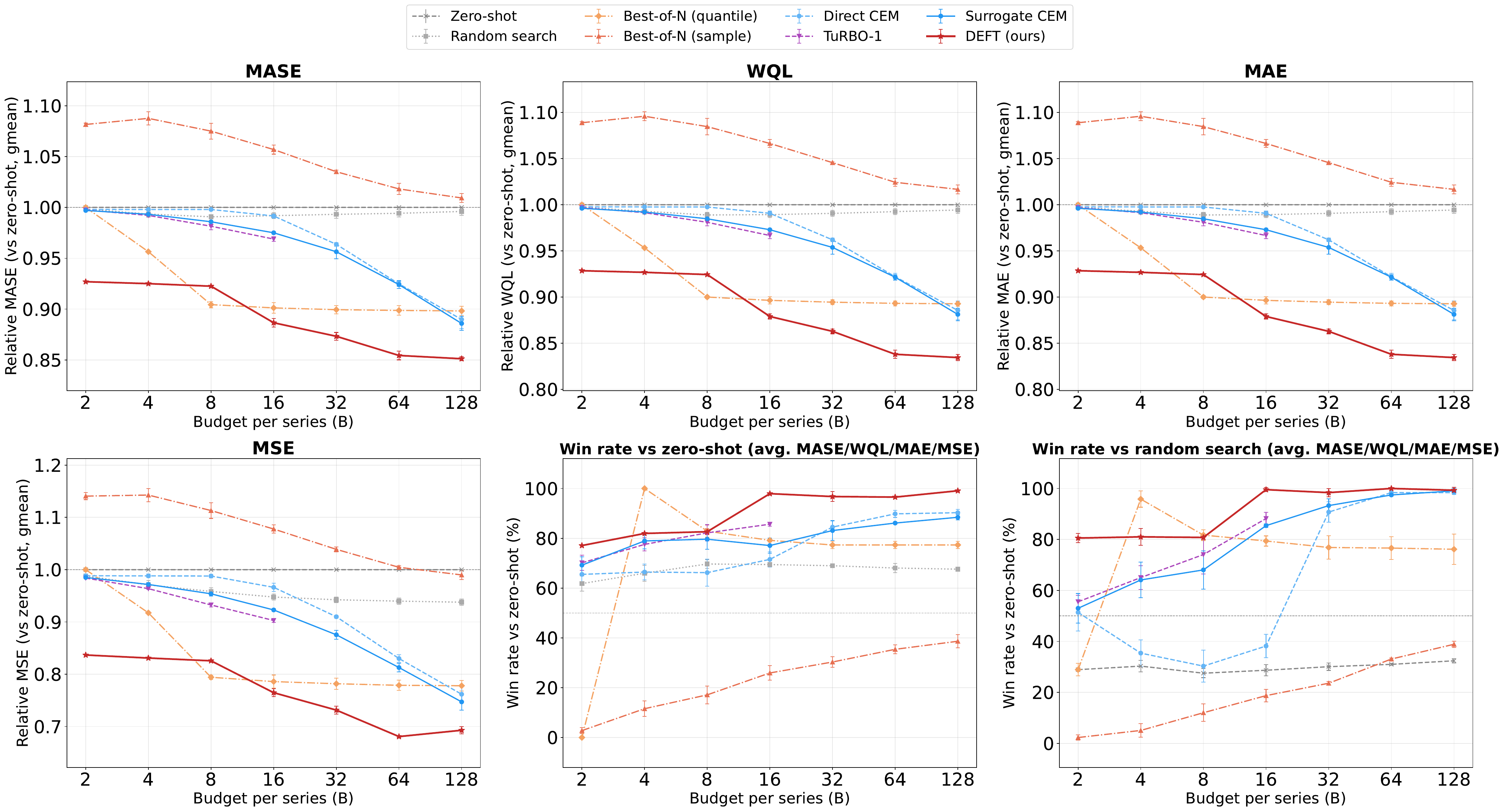}
    \caption{Performance across expert-query budgets on the GIFT-Eval Benchmark using the Moirai foundation model with Pairwise-Best expert feedback.}
    \label{fig:budget-giftevalbm-moirai-pairwisebest-noise0}
\end{figure*}

For the Moirai foundation model, we report the performance curves across expert-query budgets for different expert-feedback settings as follows:

\begin{itemize}
    \item Figure~\ref{fig:budget-chronosbm-moirai-rating3-noise0}: \textbf{Benchmark:} Chronos Benchmark; \textbf{Foundation model:} Moirai; \textbf{Expert feedback:} 3-Level Rating.

    \item Figure~\ref{fig:budget-chronosbm-moirai-rating5-noise0}: \textbf{Benchmark:} Chronos Benchmark; \textbf{Foundation model:} Moirai; \textbf{Expert feedback:} 5-Level Rating.

    \item Figure~\ref{fig:budget-chronosbm-moirai-pairwise-noise0}: \textbf{Benchmark:} Chronos Benchmark; \textbf{Foundation model:} Moirai; \textbf{Expert feedback:} Pairwise.

    \item Figure~\ref{fig:budget-chronosbm-moirai-pairwisebest-noise0}: \textbf{Benchmark:} Chronos Benchmark; \textbf{Foundation model:} Moirai; \textbf{Expert feedback:} Pairwise-Best.


    \item Figure~\ref{fig:budget-giftevalbm-moirai-rating3-noise0}: \textbf{Benchmark:} GIFT-Eval Benchmark; \textbf{Foundation model:} Moirai; \textbf{Expert feedback:} 3-Level Rating.

    \item Figure~\ref{fig:budget-giftevalbm-moirai-rating5-noise0}: \textbf{Benchmark:} GIFT-Eval Benchmark; \textbf{Foundation model:} Moirai; \textbf{Expert feedback:} 5-Level Rating.

    \item Figure~\ref{fig:budget-giftevalbm-moirai-pairwise-noise0}: \textbf{Benchmark:} GIFT-Eval Benchmark; \textbf{Foundation model:} Moirai; \textbf{Expert feedback:} Pairwise.

    \item Figure~\ref{fig:budget-giftevalbm-moirai-pairwisebest-noise0}: \textbf{Benchmark:} GIFT-Eval Benchmark; \textbf{Foundation model:} Moirai; \textbf{Expert feedback:} Pairwise-Best.

\end{itemize}

\pagebreak

\begin{algorithm}[t]
\caption{Random Search}
\label{alg:random-search}
\begin{algorithmic}[1]
\Require Initial forecast $\hat{y}^{0}$, scorer $\operatorname{score(\cdot)}$, query budget $B$, scale $\sigma$
\Ensure Edited forecast $\hat{y}$

\State $\mathcal{Q}
=
\{y_i\}_{i=1}^{B},
\quad
y_i=\hat{y}^{0}+\epsilon_i,
\quad
\epsilon_i\sim\mathcal{N}(0,\sigma^2 I)$
\Comment{$B$ perturbed forecasts}

\State Include unchanged forecast $\hat{y}^{0}\in\mathcal{Q}$

\State $v_y\leftarrow \operatorname{score}(y),\quad \forall y\in\mathcal{Q}$
\Comment{Query expert}

\State $\hat{y}\leftarrow
\arg\max_{y\in\mathcal{Q}} v_y$

\State \Return $\hat{y}$

\end{algorithmic}
\end{algorithm}

\begin{algorithm}[t]
\caption{Best-of-$N$ Quantile}
\label{alg:bon-quantile}
\begin{algorithmic}[1]
\Require Foundation-model quantile head $F_\theta(\cdot\mid q)$, scorer $\operatorname{score}(\cdot)$, query budget $N$
\Ensure Edited forecast $\hat{y}$
\State Build symmetric quantile levels $\mathcal{L}=\{q_1,\dots,q_N\}\subset(0,1)$ centered at $0.5$
\Comment{$N$ evenly spaced levels, $q_{(N+1)/2}=0.5$}
\State $\mathcal{Q}=\{y_q\}_{q\in\mathcal{L}},\quad y_q\leftarrow F_\theta(\cdot\mid q)$
\Comment{$N$ deterministic full-horizon trajectories}
\State $v_y\leftarrow \operatorname{score}(y),\quad \forall y\in\mathcal{Q}$
\Comment{Query expert}
\State $\hat{y}\leftarrow\arg\max_{y\in\mathcal{Q}} v_y$
\State \Return $\hat{y}$
\end{algorithmic}
\end{algorithm}

\begin{algorithm}[h]
\caption{Best-of-$N$ Sample}
\label{alg:bon-sample}
\begin{algorithmic}[1]
\Require Foundation model with sampler $\operatorname{Sample}_\theta(\cdot)$ or quantile head $F_\theta(\cdot\mid q)$, scorer $\operatorname{score}(\cdot)$, query budget $N$
\Ensure Edited forecast $\hat{y}$
\If{model supports native sampling}
    \State $\mathcal{Q}=\{y_i\}_{i=1}^{N-1},\quad y_i\sim \operatorname{Sample}_\theta(\cdot)$
    \Comment{$N{-}1$ stochastic trajectories}
\Else
    \State Fix dense quantile grid $\mathcal{G}\subset(0,1)$
    \For{$i=1,\dots,N-1$}
        \State $u_{i,t}\sim\mathcal{U}(0,1),\quad \forall t=1,\dots,H$
        \Comment{i.i.d.\ per-timestep uniform draw}
        \State $y_i[t]\leftarrow F_\theta^{-1}(u_{i,t}\mid t,\mathcal{G}),\quad \forall t$
        \Comment{per-timestep inverse-CDF sampling}
    \EndFor
    \State $\mathcal{Q}=\{y_i\}_{i=1}^{N-1}$
\EndIf
\State Include median forecast $y_{0.5}\leftarrow F_\theta(\cdot\mid 0.5)\in\mathcal{Q}$
\Comment{always included}
\State $v_y\leftarrow \operatorname{score}(y),\quad \forall y\in\mathcal{Q}$
\Comment{Query expert}
\State $\hat{y}\leftarrow\arg\max_{y\in\mathcal{Q}} v_y$
\State \Return $\hat{y}$
\end{algorithmic}
\end{algorithm}

\begin{algorithm}[h]
\caption{Direct CEM}
\label{alg:direct-cem}
\begin{algorithmic}[1]
\Require Initial forecast $\hat{y}^{0}$, scorer $\operatorname{score}(\cdot)$, query budget $B$, per-round size $b$, elite frac $\rho$
\Ensure Edited forecast $\hat{y}$
\State $\mu\leftarrow \hat{y}^{0}$, initialize $\sigma$, $\mathcal{A}\leftarrow\emptyset$, $q\leftarrow0$, $r\leftarrow0$
\State $n_e \leftarrow \max\!\big(1,\ \operatorname{round}(\rho b)\big)$
\Comment{elite count, fixed}
\While{$q < B$}
    \State $r\leftarrow r+1$;\quad $b_r \leftarrow \min(b,\, B-q)$
    \Comment{partial last round keeps spend exact}
    \State $\mathcal{Q}_{r}=\{\mu\}\cup\{y_i\}_{i=1}^{b_r-1},
    \quad y_i \sim \mathcal{N}(\mu,\operatorname{diag}(\sigma^2))$
    \Comment{mean re-queried $+$ $b_r{-}1$ samples}
    \State $v_y\leftarrow \operatorname{score}(y),\quad \forall y\in\mathcal{Q}_{r}$
    \Comment{Query expert}
    \State $\mathcal{A}\leftarrow \mathcal{A}\cup\{(y,v_y):y\in\mathcal{Q}_{r}\}$
    \State $\mathcal{E}_{r}\leftarrow \operatorname{TopK}_{\min(n_e,\,b_r)}
    \{y\in\mathcal{Q}_{r}:v_y\}$
    \Comment{elites from \emph{queried} set}
    \State $\mu\leftarrow \operatorname{Mean}(\mathcal{E}_{r})$;\quad
    $\sigma\leftarrow \max\!\big(\operatorname{Std}(\mathcal{E}_{r}),\,\sigma_{\min}\big)$
    \Comment{refit, floored at $\sigma_{\min}$}
    \State $q\leftarrow q+b_r$
\EndWhile
\State $\hat{y}\leftarrow\arg\max_{y\in\mathcal{A}} v_y$
\State \Return $\hat{y}$
\end{algorithmic}
\end{algorithm}















\begin{algorithm}[h]
\caption{Surrogate CEM}
\label{alg:surrogate-cem}
\begin{algorithmic}[1]
\Require Initial forecast $\hat{y}^{0}$, scorer $\operatorname{score}(\cdot)$, query budget $B$, per-round size $b$, pool size $N\!\gg\!b$, elite frac $\rho$
\Ensure Edited forecast $\hat{y}$
\State $\mu\leftarrow \hat{y}^{0}$, initialize $\sigma$, $\mathcal{A}\leftarrow\emptyset$, $q\leftarrow0$, $r\leftarrow0$, $S_{\phi}\leftarrow\varnothing$
\State $n_e \leftarrow \max\!\big(1,\ \operatorname{round}(\rho N)\big)$
\Comment{elite count over the pool}
\While{$q < B$}
    \State $r\leftarrow r+1$;\quad $b_r \leftarrow \min(b,\, B-q)$
    \Comment{partial last round keeps spend exact}
    \State $\mathcal{Y}_{r}=\{y_i\}_{i=1}^{N},\quad y_i\sim\mathcal{N}(\mu,\operatorname{diag}(\sigma^2))$
    \Comment{pool, surrogate-scored only (no expert queries)}
    \If{$S_{\phi}\neq\varnothing$}
        \State $\tilde{s}_y\leftarrow S_{\phi}(y)\quad \forall y\in\mathcal{Y}_{r}$
        \Comment{cheap surrogate scoring}
        \State $\mathcal{Q}_{r}\leftarrow\{\mu\}\cup\operatorname{Select}_{b_r-1}(\mathcal{Y}_{r},\tilde{s},\mathcal{A})$
        \Comment{exploit top-$\tilde s$, explore far from $\mathcal{A}$}
    \Else
        \State $\mathcal{Q}_{r}\leftarrow\{\mu\}\cup\{b_r{-}1\text{ uniform-random from }\mathcal{Y}_{r}\}$
        \Comment{round~1, no surrogate yet}
    \EndIf
    \State $v_y\leftarrow \operatorname{score}(y)\quad \forall y\in\mathcal{Q}_{r}$
    \Comment{query expert; $|\mathcal{Q}_r|{=}b_r$}
    \State $\mathcal{A}\leftarrow\mathcal{A}\cup\{(y,v_y):y\in\mathcal{Q}_{r}\}$
    \State $S_{\phi}\leftarrow\operatorname{Fit}(\mathcal{A})$
    \Comment{refit \emph{after} querying}
    \State $\tilde{s}_y\leftarrow S_{\phi}(y)\quad \forall y\in\mathcal{Y}_{r}$
    \Comment{surrogate-score the whole pool}
    \State $\tilde{s}_y\leftarrow v_y\quad \forall y\in\mathcal{Q}_{r}$
    \Comment{overwrite queried subset ($\mathcal{Q}_r\!\subseteq\!\mathcal{Y}_r$) with exact scores}
    \State $\mathcal{E}_{r}\leftarrow\operatorname{TopK}_{n_e}\{y\in\mathcal{Y}_{r}:\tilde{s}_y\}$
    \Comment{elites over \emph{full} pool}
    \State $\mu\leftarrow \operatorname{Mean}(\mathcal{E}_{r})$;\quad
    $\sigma\leftarrow \max\!\big(\operatorname{Std}(\mathcal{E}_{r}),\,\sigma_{\min}\big)$
    \Comment{refit, floored at $\sigma_{\min}$}
    \State $q\leftarrow q+b_r$
\EndWhile
\State $\hat{y}\leftarrow\arg\max_{y\in\mathcal{A}} v_y$
\State \Return $\hat{y}$
\end{algorithmic}
\end{algorithm}

\begin{algorithm}[h]
\caption{TuRBO-1}
\label{alg:turbo}
\begin{algorithmic}[1]
\Require Initial forecast $\hat{y}^{0}$, scorer $\operatorname{score}(\cdot)$, query budget $B$, scale $\sigma$
\Ensure Edited forecast $\hat{y}$

\State Define $y(x)=\hat{y}^{0}+(2x-1)\sigma,\quad x\in[0,1]^H$

\State Initialize $\mathcal{X}$ with Sobol samples including $x_0=\frac{1}{2}\mathbf{1}$

\State $s_x\leftarrow \operatorname{score}(y(x)),\quad \forall x\in\mathcal{X}$
\Comment{Initial expert queries}

\State $\mathcal{A}\leftarrow\{(x,s_x):x\in\mathcal{X}\}$, $q\leftarrow|\mathcal{X}|$

\State $x^\star\leftarrow\arg\max_{(x,s)\in\mathcal{A}}s$, initialize trust-region length $\ell$

\While{$q<B$}

    \State Fit GP surrogate $g$ on $\mathcal{A}$

    \State Construct trust region
    $\mathcal{T}=\operatorname{TR}(x^\star,\ell)\cap[0,1]^H$

    \State Draw candidate set $\mathcal{C}$ from $\mathcal{T}$ using Sobol samples

    \State Sample $\tilde{g}(x)$ from GP posterior for all $x\in\mathcal{C}$

    \State $x_{\mathrm{next}}\leftarrow\arg\max_{x\in\mathcal{C}}\tilde{g}(x)$

    \State $s_{\mathrm{next}}\leftarrow \operatorname{score}(y(x_{\mathrm{next}}))$
    \Comment{Query expert}

    \State $\mathcal{A}\leftarrow\mathcal{A}\cup\{(x_{\mathrm{next}},s_{\mathrm{next}})\}$

    \If{$s_{\mathrm{next}}>s_{x^\star}$}
        \State $x^\star\leftarrow x_{\mathrm{next}}$
        \State Increase $\ell$
    \Else
        \State Decrease $\ell$ after repeated failures
    \EndIf

    \If{$\ell<\ell_{\min}$}
        \State Restart trust region and reset $\ell$
    \EndIf

    \State $q\leftarrow q+1$

\EndWhile

\State $\hat{y}\leftarrow y(x^\star)$

\State \Return $\hat{y}$

\end{algorithmic}
\end{algorithm}

\begin{algorithm}[h]
\caption{DEFT: Decomposed Expert-Guided ForecasT}
\label{alg:deft}
\begin{algorithmic}[1]
\Require Initial forecast $\hat{y}^{0}$, foundation-model samples $\mathcal{P}$, scorer $\operatorname{score}(\cdot)$, query budget $B$, seed budget $B_0$, per-round budget $b$, $n_r$ search rounds ($B=B_0+b\times n_r$), elite fraction $\rho$, component counts $n_T,n_S$
\Ensure Edited forecast $\hat{y}$

\Statex \textbf{Stage 1: decomposed Seed-Pool Scan}
\State $\mathcal{P}_T\leftarrow\{\operatorname{MA}_{w}(y):y\in\mathcal{P}\}$; 
$\mathcal{P}_S\leftarrow\{y-\operatorname{MA}_{w}(y):y\in\mathcal{P}\}$
\Comment{Decompose TFM samples}
\State Select $K$ paired components $\{(T_i,S_i)\}_{i=1}^{K}$ from $(\mathcal{P}_T,\mathcal{P}_S)$

\State Form $B_0$ recombinations
$
\mathcal{Q}_0=\{T_i+S_j:(i,j)\in\mathcal{I}_0\}
$

\State $v_{ij}\leftarrow \operatorname{score}(T_i+S_j),\quad (i,j)\in\mathcal{I}_0$
\Comment{Query expert on prior recombinations}

\State $u_i^T\leftarrow \max_{j:(i,j)\in\mathcal{I}_0}v_{ij}$; 
$u_j^S\leftarrow \max_{i:(i,j)\in\mathcal{I}_0}v_{ij}$
\Comment{Reuse each score for both components}

\State $\mathcal{E}_0^T\leftarrow\operatorname{TopK}_{\rho K}\{T_i:u_i^T\}$; 
$\mathcal{E}_0^S\leftarrow\operatorname{TopK}_{\rho K}\{S_j:u_j^S\}$

\State $\mu_T\leftarrow\operatorname{Mean}(\mathcal{E}_0^T)$; 
$\mu_S\leftarrow\operatorname{Mean}(\mathcal{E}_0^S)$

\State $\sigma_T\leftarrow\operatorname{Std}(\mathcal{P}_T)$; 
$\sigma_S\leftarrow\operatorname{Std}(\mathcal{P}_S)$
\Comment{Initialize search scale from model uncertainty}

\State $\hat{y}\leftarrow\arg\max_{(i,j)\in\mathcal{I}_0}v_{ij}$; 
$q\leftarrow B_0$

\Statex {\textbf{Stage 2: decomposed CEM search}}

\For{$r=1,\ldots,n_r$}

    \State $\mathcal{T}_r=\{T_i\}_{i=1}^{n_T},
    \quad
    T_i\sim\mathcal{N}(\mu_T,\operatorname{diag}(\sigma_T^2))$

    \State $\mathcal{S}_r=\{S_j\}_{j=1}^{n_S},
    \quad
    S_j\sim\mathcal{N}(\mu_S,\operatorname{diag}(\sigma_S^2))$

    \State Form $b$ recombinations
    $
    \mathcal{Q}_r=\{T_i+S_j:(i,j)\in\mathcal{I}_r\}
    $
    \Comment{$\mathcal{I}_r$ covers sampled components}

    \State $v_{ij}\leftarrow \operatorname{score}(T_i+S_j),
    \quad (i,j)\in\mathcal{I}_r$
    \Comment{Query expert}

    \State $u_i^T\leftarrow
    \max_{j:(i,j)\in\mathcal{I}_r}v_{ij}$; 
    $u_j^S\leftarrow
    \max_{i:(i,j)\in\mathcal{I}_r}v_{ij}$
    \Comment{Component-level utility}

    \State $\mathcal{E}_r^T\leftarrow
    \operatorname{TopK}_{\rho n_T}\{T_i:u_i^T\}$; 
    $\mathcal{E}_r^S\leftarrow
    \operatorname{TopK}_{\rho n_S}\{S_j:u_j^S\}$

    \State $\mu_T\leftarrow\operatorname{Mean}(\mathcal{E}_r^T)$; 
    $\sigma_T\leftarrow\operatorname{Std}(\mathcal{E}_r^T)$

    \State $\mu_S\leftarrow\operatorname{Mean}(\mathcal{E}_r^S)$; 
    $\sigma_S\leftarrow\operatorname{Std}(\mathcal{E}_r^S)$
    \Comment{Refit component proposals}

    \State $\hat{y}\leftarrow
    \arg\max\{\hat{y},\,T_i+S_j:(i,j)\in\mathcal{I}_r\}$ under $\operatorname{score}(\cdot)$

    \State $q\leftarrow q+b$

    \If{$q\ge B$}
        \State \textbf{break}
    \EndIf

\EndFor

\State \Return $\hat{y}$
\end{algorithmic}
\end{algorithm}

\begin{algorithm}[t]
\caption{DEFT budget allocation}
\label{alg:deft-budget}
\begin{algorithmic}[1]
\Require Total expert-query budget $B$; trend fraction $\tau$ (default $0.5$)
\Ensure Seed-pool size $B_0$, per-round queries $b$, rounds $n_r$; per-round split $n_T,n_S,K$
\State $s \gets \max(2,\ \lfloor B/2 \rfloor)$ \Comment{split budget: pool vs.\ CEM refinement}
\State $b \gets \max\!\big(4,\ \lfloor B/8 \rfloor\big)$ \Comment{queries per round, floored at $4$}
\State $n_r \gets \min\!\big(4,\ \lfloor s/b \rfloor\big)$ \Comment{refinement rounds, capped at $4$}
\If{$n_r = 0$ \textbf{and} $B \ge b+2$}
  \State $n_r \gets 1$ \Comment{ensure one round once affordable}
\EndIf
\If{$B \le 16$}
  \State $n_r \gets \min(n_r,\,1)$ \Comment{small budget: at most one round}
\EndIf
\State $B_0 \gets \max\!\big(2,\ B - b\cdot n_r\big)$ \Comment{remaining budget scans the seed pool}
\State $n_T \gets \max\!\big(2,\ \mathrm{round}(\tau\, b)\big)$;\quad $n_S \gets \max\!\big(2,\ b - n_T\big)$
\State $K \gets \max\!\big(2,\ \lceil \sqrt{B_0}\,\rceil\big)$ \Comment{$K$ trend \& $K$ seasonal seeds, paired $1{:}1$}
\State \Return $B_0,\ b,\ n_r,\ n_T,\ n_S,\ K$
\end{algorithmic}
\end{algorithm}

\pagebreak
\section{Full Ablation Results for DEFT Design Choices}
\label{appendix:ablation}
\subsection{Variant Descriptions}
\label{appendix:ablation-variants}

All ablation variants share the same evaluation protocol as the main experiments. Each variant modifies exactly one design choice of DEFT while keeping everything else fixed. Table~\ref{tab:ablation-variants} summarises the variants; we describe each in detail below.

\begin{table}[h]
\centering
\caption{Summary of ablation variants. Ticks indicate which components are active.}
\label{tab:ablation-variants}
\resizebox{\linewidth}{!}{%
\small
\begin{tabular}{lcccc}
\toprule
\textbf{Variant} & \textbf{Seed pool} & \textbf{T/S decomp.} & \textbf{CEM rounds} & \textbf{Pool mode} \\
\midrule
Zero shot median     & -- & -- & -- & -- \\
Random search        & -- & -- & \checkmark (Gaussian) & -- \\
Pool only            & \checkmark & \checkmark & -- & max \\
No seed pool         & -- & \checkmark & \checkmark & -- \\
CEM + prior          & \checkmark & -- & \checkmark & max \\
Mean marginal        & \checkmark & \checkmark & \checkmark & mean \\
Trend only           & \checkmark & \checkmark (trend) & \checkmark & max \\
Seasonal only        & \checkmark & \checkmark (seasonal) & \checkmark & max \\
\textbf{DEFT (full)} & \checkmark & \checkmark & \checkmark & max \\
\bottomrule
\end{tabular}}
\end{table}

\paragraph{Zero shot median.}
The foundation model's median forecast with no expert queries and no optimisation. This is the lower-bound baseline that DEFT improves upon.

\paragraph{Random search.}
At each query, a Gaussian perturbation is applied to the current median forecast across the full trajectory, and the expert selects the best candidate. This is the primary reference baseline: it matches DEFT's query budget but uses unstructured random exploration with no pool and no decomposition.

\paragraph{Pool only.}
The full quantile-prior seed pool is scanned and the best candidate is selected, but no Gaussian CEM rounds follow ($n_\text{rounds}=0$). The pool size equals the full budget $B$. This variant isolates the gain from the seed pool alone, without any iterative refinement.

\paragraph{No seed pool.}
The pre-CEM seed pool scan is skipped entirely; the full budget is allocated to Gaussian CEM rounds starting from the zero-shot median. Trend--seasonal decomposition is retained. This variant isolates the contribution of the seed pool by comparing against a CEM-only baseline that has the same number of refinement rounds but no informed initialisation.

\paragraph{CEM + prior.}
The seed pool is used for initialisation, and CEM is run on the \emph{full trajectory} without trend--seasonal decomposition. This variant isolates the contribution of decomposition: comparing \textsc{CEM-with-Prior} to full DEFT shows how much is gained by structuring the search space into trend and seasonal components rather than optimising the raw trajectory directly.

\paragraph{Mean marginal.}
Identical to full DEFT except that the component utility scores are computed as the mean
over paired expert scores rather than the maximum:
\begin{equation}
u_i^T = \frac{1}{|\{j:\,(i,j)\in\mathcal{I}_0\}|}\sum_{j:\,(i,j)\in\mathcal{I}_0} v_{ij},
\qquad
u_j^S = \frac{1}{|\{i:\,(i,j)\in\mathcal{I}_0\}|}\sum_{i:\,(i,j)\in\mathcal{I}_0} v_{ij},
\end{equation}
in place of the max-pool utility in Eq.~\ref{eq:utility}.
This tests whether the specific form of score propagation across components affects sample efficiency.

\paragraph{Trend only.}
The seasonal component is frozen at the moving-average seasonal extracted from the zero-shot median; only the trend component is searched via CEM. This isolates the contribution of trend editing.

\paragraph{Seasonal only.}
The trend component is frozen at the moving-average trend extracted from the zero-shot median; only the seasonal component is searched via CEM. This isolates the contribution of seasonal editing.

\subsection{Results for Moirai}
\label{appendix:ablation-results}

\begin{figure}[h]
    \centering
    \includegraphics[width=\linewidth]{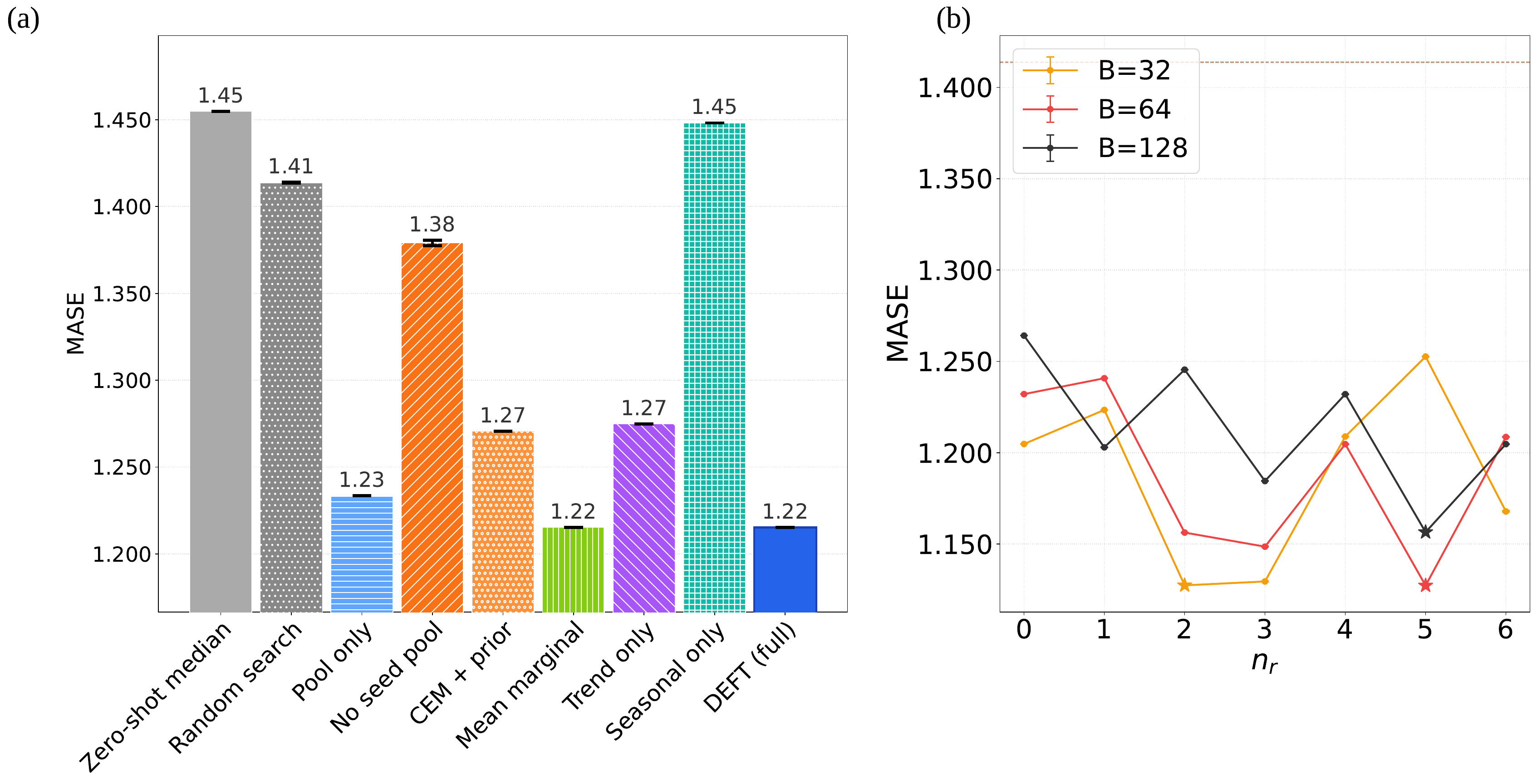}
      \caption{Ablation study of DEFT on the Moirai backbone under budgets $B \in \{32,64,128\}$.
    \textbf{(a)} Mean MASE per ablation variant at each budget.
    \textbf{(b)} Rounds sweep: mean MASE as a function of the number of decomposed CEM rounds $n_{\text{rounds}}$ at fixed budget (with $n_{\text{prior}}$ shrinking correspondingly); dashed lines indicate the random-search baseline for each budget, and $\bigstar$ marks the optimal $n_{\text{rounds}}$ for each budget.}
  
    \label{fig:app:ablation-moirai}
\end{figure}

Figures~\ref{fig:app:ablation-moirai} reproduces the ablation plots of Figure~\ref{fig:ablation-timesfm-chronos} for the Moirai backbone. The qualitative ordering of variants is consistent across all three backbones. DEFT (full) achieves the lowest MASE, the pool-only scan is highly competitive at small budgets, and \textsc{Seasonal-Only} is consistently the weakest decomposition variant.


\begin{table}[t]
\centering
\caption{Inference time (wall-clock seconds per series). Every method issues the same number of expert queries ($B$) and a single
foundation-model forward pass; they differ only in search overhead. \textsc{DEFT}
tracks best-of-$N$ and stays well below CEM, and remains feasible on the
high-dimensional MD state where surrogate CEM runs out of memory.}
\label{tab:compute-cost}
\setlength{\tabcolsep}{6pt}
\begin{tabular}{@{}lcccc@{}}
\toprule
& \multicolumn{2}{c}{\textbf{MD}} & \multicolumn{2}{c}{\textbf{Main}} \\
\cmidrule(lr){2-3}\cmidrule(lr){4-5}
Method & $B{=}32$ & $B{=}128$ & $B{=}32$ & $B{=}128$ \\
\midrule
Zero-shot      & 0.011 & 0.013 & 0.000 & 0.000 \\
Best-of-$N$    & 0.586 & 3.162 & 0.000 & 0.000 \\
Random search  & 1.018 & 4.441 & 0.000 & 0.000 \\
CEM            & 3.862 & 10.246 & 0.002 & 0.004 \\
Surrogate CEM  & \textcolor{red}{OOM} & \textcolor{red}{OOM} & 0.020 & 0.043 \\
\rowcolor{deftblue}
\textsc{DEFT}  & 0.827 & 4.395 & 0.002 & 0.005 \\
\bottomrule
\end{tabular}
\end{table}

\end{document}